\documentclass[lettersize,journal]{IEEEtran}
\usepackage{array}
\usepackage[caption=false,font=normalsize,labelfont=sf,textfont=sf]{subfig}
\usepackage{textcomp}
\usepackage{stfloats}
\usepackage{booktabs} 
\usepackage{amsthm}

\usepackage[utf8]{inputenc}
\usepackage[T1]{fontenc}
\usepackage{tabularx} 
\usepackage{array}

\usepackage{url}
\usepackage{verbatim}
\usepackage{graphicx}
\def\BibTeX{{\rm B\kern-.05em{\sc i\kern-.025em b}\kern-.08em
    T\kern-.1667em\lower.7ex\hbox{E}\kern-.125emX}}
\usepackage{balance}
\usepackage{algorithm}       
\usepackage{algpseudocode}    
\usepackage{amsmath, amssymb, amsfonts}
\usepackage{graphicx}        
\usepackage{subcaption}      
\usepackage{placeins}
\usepackage{multirow}
\usepackage{xcolor}
\usepackage{listings}
\usepackage{url}
\usepackage{hyperref} 
\usepackage{enumitem}
\definecolor{lightcream}{rgb}{1.0, 1.0, 0.96}
\lstdefinelanguage{MyC++}{
  language=C++,
  morekeywords=[1]{function,Graph,container,int,totalEpoch,Graph,gnn},
  morekeywords=[2]{initializeLayers,forwardPass,backPropogation,optimizer},
  keywordstyle=[1]\color{blue}\bfseries,
  keywordstyle=[2]\color{teal}\bfseries,
  commentstyle=\color{gray}\itshape,
  stringstyle=\color{red},
}

\begin{document}
\title{Morphling: Fast, Fused, and Flexible GNN Training at Scale}

\author{
    \begin{tabular}{cc}
        \Large Anubhab & \Large Rupesh Nasre \\
        \large \href{mailto:anubhavkhajuria5@gmail.com}{anubhavkhajuria5@gmail.com} & \large \href{mailto:rupesh@cse.iitm.ac.in}{rupesh@cse.iitm.ac.in} \\
        \large \textit{IIT Madras}, India & \large \textit{IIT Madras}, India
    \end{tabular}
}

\maketitle

\begin{abstract}
Graph Neural Networks (GNNs) present a fundamental hardware challenge by fusing irregular, memory-bound graph traversals with regular, compute-intensive dense matrix operations. While frameworks such as PyTorch Geometric (PyG) and Deep Graph Library (DGL) prioritize high-level usability, they fail to address these divergent execution characteristics. As a result, they rely on generic kernels that suffer from poor cache locality, excessive memory movement, and substantial intermediate allocations. To address these limitations, we present Morphling, a domain-specific code synthesizer designed to bridge this gap. Morphling compiles high-level GNN specifications into portable, backend-specialized implementations targeting OpenMP, CUDA, and MPI. It achieves this by instantiating a library of optimized, architecture-aware primitives tailored to each execution environment. Morphling also incorporates a runtime sparsity-aware execution engine that dynamically selects dense or sparse execution paths using input feature statistics, reducing unnecessary computation on zero-valued entries. We evaluate Morphling on eleven real-world datasets spanning diverse graph structures, feature dimensionalities, and sparsity regimes. The results show that Morphling improves per-epoch training throughput by an average of 20$\times$ on CPUs, 19$\times$ on GPUs, and 6$\times$ in distributed settings over PyG and DGL, with peak speedups reaching 66$\times$. Morphling’s memory-efficient layouts further reduce peak memory consumption by up to 15$\times$, enabling large-scale GNN training on commodity hardware. These findings demonstrate that specialized, architecture-aware code synthesis provides an effective and scalable path toward high-performance GNN execution across diverse parallel and distributed platforms.

\end{abstract}


\section{Introduction}
GNNs have become the standard for learning on graph-structured data, powering applications ranging from recommendation systems to scientific computing. Unlike traditional neural networks that treat data points in isolation, GNNs generate topology-aware representations by leveraging the underlying structure of the data. They achieve this through iterative message passing, where nodes update their embeddings by aggregating high-dimensional features from their local neighborhoods. Their effectiveness stems from integrating irregular, memory-bound graph traversals directly with regular, compute-intensive dense matrix operations. This sparse--dense computational hybrid poses unique systems challenges that are poorly addressed by existing general-purpose frameworks.

To manage this complexity, practitioners typically turn to frameworks like PyTorch Geometric (PyG)~\cite{pyg} and Deep Graph Library (DGL)~\cite{dgl}, which prioritize high-level programmability and seamless integration with PyTorch’s autograd engine. However, because these frameworks represent node features as dense tensors, they are forced to materialize substantial intermediate data during message passing and to execute redundant operations on naturally sparse inputs~\cite{dgcl}. These choices introduce excessive memory movement, poor cache locality, and costly intermediate allocations. These challenges intensify significantly on graphs with large feature dimensions or millions of edges. Moreover, PyG and DGL lack unified abstractions for efficient execution across heterogeneous hardware. While CPUs thrive on cache-aware tiling and SIMD vectorization, GPUs require warp-synchronous kernels with coalesced memory access, and distributed clusters necessitate rigorous communication–computation overlap. Since existing frameworks abstract away these platform-specific decisions, they face fundamental challenges in ensuring performance portability across architectures.

This optimization gap places practitioners in a difficult bind: they must choose between the development speed of high-level APIs and the raw performance of hand-tuned execution. Opting for performance requires manually implementing specialized kernels for CPUs, GPUs, and distributed clusters; however, maintaining these distinct codebases is error-prone and becomes increasingly unsustainable as hardware architectures diversify. While specialized systems like GNNAdvisor~\cite{gnnadvisor} and DGCL~\cite{dgcl} attempt to bridge this gap, they function as single-platform systems that lack cross-platform portability and leave the overheads of dense feature processing unresolved. Alternatively, graph DSLs such as StarPlat~\cite{starplat}, GraphIt~\cite{graphit}, and Ligra~\cite{ligra} generate high-performance code but are limited to classical graph analytics, lacking essential support for backpropagation, optimizer updates, or feature sparsity. Similarly, tensor compilers such as Halide~\cite{halide} and TVM~\cite{tvm} provide powerful scheduling abstractions but treat GNNs as generic tensor programs. Because they lack domain-specific primitives, they cannot perform graph-aware optimizations across the full training loop. These gaps highlight the need for a domain-specific approach that captures GNN computation at a high level while enabling backend-specialized code generation. The solution requires a sparsity-aware, memory-efficient architecture that scales transparently across diverse backends.

To bridge this gap, we present Morphling, a synthesis-oriented framework that generates backend-specialized C++ training code from compact high-level GNN specifications. Through synthesis, Morphling integrates architecture-aware kernels for CPUs, GPUs, and distributed systems, chooses between dense and sparse execution paths based on input feature statistics, and embeds distributed communication and partitioning logic directly into the generated code. This design enables execution plans that align closely with each backend’s memory hierarchy and parallelism model while preserving the simplicity of high-level GNN programming.

The contributions of this paper are as follows:
\begin{itemize}

    \item \textbf{A GNN-native domain-specific language} built on the StarPlat DSL and extended with GNN-specific operators, supporting forward propagation, neighborhood aggregation, backpropagation, and optimizer updates while abstracting away backend-specific complexity.

    \item \textbf{A multi-backend code synthesis framework} that generates optimized OpenMP, CUDA, and MPI implementations from a single high-level program, including cache-tiled CPU kernels, warp-synchronous GPU aggregation kernels, and communication-overlapped distributed execution.

    \item \textbf{A runtime sparsity-aware execution engine} that automatically selects efficient dense or sparse feature-processing paths based on feature statistics, exploiting feature-level sparsity without changing model semantics.

    \item \textbf{A comprehensive experimental evaluation} across eleven real-world datasets demonstrating significant throughput improvements over PyG and DGL on CPUs, GPUs, and distributed clusters, along with up to $15\times$ lower peak memory consumption.

\end{itemize}

\section{Related Work}

\noindent\textbf{GNN Frameworks.} PyG~\cite{pyg} and DGL~\cite{dgl} dominate GNN development by offering extensive model libraries and seamless integration with PyTorch's autograd engine. However, their reliance on the gather-scatter paradigm~\cite{dgcl} limits opportunities for kernel fusion and data reuse. First, they treat node features as dense tensors, forcing the materialization of large intermediate structures and precluding optimizations for intrinsically sparse features (e.g., one-hot encodings). Second, they lack a unified intermediate representation preventing optimizations from generalizing across backends. Consequently, targeting CPUs, GPUs, and clusters necessitates managing disjoint optimization strategies, impeding true performance portability. Some of these limitations have motivated several systems that focus exclusively on GPU execution.

Beyond general-purpose frameworks, some systems focus exclusively on GPU execution. These systems improve GNN throughput by employing custom GPU kernels with hardware-aware optimizations such as 2D partitioning and neighbor grouping, as demonstrated by GNNAdvisor~\cite{gnnadvisor} and DGCL~\cite{dgcl}. However, like PyG and DGL, they assume dense feature tensors, which limits the applicability of their 
optimizations to workloads with intrinsic feature sparsity. While they achieve high single-device throughput, they are tightly coupled to GPU architectures and thus lack the flexibility to generate code for CPUs or MPI-based clusters.

In the distributed setting, industrial systems like AliGraph~\cite{aligraph}, Euler~\cite{euler}, and DistDGL~\cite{distdgl} target billion-scale graph training. However, their primary optimizations focus on distributed storage, I/O minimization, and sampling strategies rather than the computational efficiency of the core GNN computation. These systems emphasize scale and data movement rather than computation efficiency, whereas Morphling targets the optimization of the core GNN kernels across CPU, GPU, and distributed backends.

\noindent\textbf{Graph Processing DSLs.}
By separating algorithm specification from execution scheduling, GraphIt~\cite{graphit} enables optimizations for a wide range of graph analytics workloads. It primarily generates optimized CPU code for classical graph algorithms (PageRank, BFS) but does not support GNN training primitives like automatic differentiation or backpropagation. Other traversal-oriented frameworks, such as Ligra~\cite{ligra}, Galois~\cite{galois}, and Gemini~\cite{gemini}, provide abstractions for parallel graph traversals, but they lack support for neural network computation and differentiation. Morphling was designed to bridge this functional gap. It extends StarPlat’s multi-backend code generation with domain-specific, primitives required for GNN training.

\noindent\textbf{Code Generation.}
General-purpose compilers such as Halide~\cite{halide} and TVM~\cite{tvm} promote performance portability by separating algorithm specification from execution scheduling. 
Halide~\cite{halide} introduced this model for image processing, and TVM~\cite{tvm} extends it to deep learning with auto-scheduling. However, TVM still treats GNNs as conventional neural networks and therefore cannot exploit graph-specific optimizations. Similarly, TACO~\cite{taco} generates efficient sparse tensor kernels but operates only at the single-operation level and does not support full training pipelines. Meanwhile, polyhedral compilers such as Tiramisu~\cite{tiramisu} and Tensor Comprehensions~\cite{tc} require affine loop structures and substantial manual tuning, making them not well aligned with the 
Irregular sparsity patterns are inherent to GNN computation.

\noindent\textbf{{Performance Portability Frameworks.}}
Performance portability aims to write code once and achieve efficient execution across diverse hardware without manual porting. Traditional approaches include:

\paragraph{Low-Level Abstractions.}
Kokkos~\cite{kokkos} and RAJA~\cite{raja} provide policy-based programming models in which developers write kernels once and specify execution policies (e.g., OpenMP or CUDA) separately.  Similarly, SYCL~\cite{sycl} and oneAPI~\cite{oneapi} provide unified C++ models for heterogeneous programming. Nonetheless, achieving high performance depends on backend-specific tuning of work-group configurations, memory layout, and kernel parameters. OpenCL~\cite{opencl} exposes similar portability–performance tradeoffs, offering broad hardware reach but requiring substantial manual optimization to match backend capabilities. Although these frameworks support multiple backends, they operate at low abstraction levels, requiring developers to manage memory layouts, parallel execution policies, and loop transformations explicitly. As a result, porting complex algorithms such as GNN training still requires substantial manual effort, particularly when expressing the sparse–dense computation patterns and multi-stage data dependencies inherent to GNN workloads.

\paragraph{Compiler-Based Approaches.}
TVM~\cite{tvm} and Halide~\cite{halide} use high-level tensor programs with auto-scheduling to generate optimized code. TVM supports multiple backends but treats GNNs as generic neural networks, missing graph-specific optimizations. TACO~\cite{taco} generates efficient sparse tensor kernels but operates on individual operations, not full training pipelines.

\paragraph{Domain-Specific Languages.}
DSLs like GraphIt~\cite{graphit} separate algorithm specification from optimization schedules, enabling automatic exploration of parallelization and locality strategies. GraphIt targets traditional graph analytics (PageRank, BFS) on CPUs but lacks neural network primitives. Ligra~\cite{ligra} and Galois~\cite{galois} provide vertex-centric abstractions but do not support multi-backend code generation.

\section{Background}
\label{sec:background}

\subsection{Graph Neural Networks}
Graph Neural Networks (GNNs) extend deep learning to graph-structured data by iteratively aggregating information from node neighborhoods. 
A GNN layer combines two computational phases: neighbor aggregation, which is sparse, irregular, and memory-bound, and feature transformation, which is dense, regular, and compute-intensive. 
Formally, the feature update for node $v$ at layer $\ell$ is

\[
h_v^{(\ell)} = \sigma\!\left(W^{(\ell)} \cdot 
    \text{AGGREGATE}^{(\ell)}\!\left(\{h_u^{(\ell-1)} : u \in \mathcal{N}(v)\}\right)\right),
\]

where $\mathcal{N}(v)$ denotes the neighbors of $v$, $W^{(\ell)}$ is a learnable weight matrix, and $\sigma$ is a nonlinear activation.

Different GNN architectures vary in their aggregation strategies. 
GCN~\cite{gcn} uses normalized mean aggregation, GraphSAGE~\cite{graphsage} introduces sampling-based variants, GIN~\cite{gin} employs sum aggregation, and GAT~\cite{gat} incorporates attention weights to modulate neighbor importance. Despite these differences, all models share the same sparse–dense computational structure, and backpropagation through this pattern amplifies the irregularity and memory bandwidth demands, creating a challenging workload for modern hardware.

\subsection{Challenges in GNN System Design}

\paragraph{Sparse-Dense Computational Hybrid.}
GNN layers combine sparse graph operations with dense matrix computations. The aggregation step involves sparse matrix-matrix multiplication (SpMM) between the adjacency matrix $A$ and feature matrix $H$, exhibiting irregular memory access patterns. The transformation step multiplies the aggregated features by weight matrix $W$, typically implemented as dense matrix multiplication (GEMM). Existing frameworks like PyG~\cite{pyg} and DGL~\cite{dgl} optimize graph sparsity but assume features are dense, wasting computation when input features are intrinsically sparse (e.g., bag-of-words, one-hot encodings).

\paragraph{Heterogeneous Hardware Deployment}
GNN workloads span a wide range of graph sizes and therefore benefit from diverse hardware platforms: multicore CPUs for moderate sized datasets, GPUs for massively parallel aggregation, and distributed clusters for datasets that exceed single-node memory. Each platform requires a distinct optimization strategy. CPUs rely on cache-aware tiling and SIMD vectorization to mitigate irregular sparsity, while GPUs depend on warp-level parallelism and coalesced memory access to sustain throughput. Distributed systems introduce an additional layer of complexity through graph partitioning, communication scheduling, and computation–communication overlap. As a result, supporting efficient GNN execution across CPUs, GPUs, and distributed systems requires reconciling fundamentally different optimization strategies within a single programming and execution model.

\paragraph{Feature Sparsity}
Many real-world graphs contain highly sparse node features. Text-based graphs rely on bag-of-words or TF–IDF (Term Frequency–Inverse Document Frequency) encodings with sparsity levels exceeding 90\%, and categorical attributes often use one-hot or multi-hot representations. Despite this, standard GNN implementations store features in dense form, leading to substantial redundant computation, unnecessary memory traffic, and poor cache utilization when multiplying by zeros. Sparse tensor libraries in PyTorch~\cite{pytorch} and TensorFlow~\cite{tensorflow} offer sparse representations, but they require manual conversion and are not integrated with GNN operations such as neighborhood aggregation or backpropagation. As a result, efficiently supporting feature sparsity requires system-level mechanisms that can adapt execution to the sparsity structure of the input features.

\subsection{StarPlat: A versatile DSL for graph analytics}

StarPlat~\cite{starplat} is a graph analytics DSL that generates optimized code for OpenMP (shared-memory CPUs), CUDA (GPUs), and MPI (distributed clusters) from a single high-level specification. The system consists of:

\paragraph{Front-End DSL.}
Users express graph algorithms using vertex-centric or edge-centric abstractions with parallel loops, reductions, and conditional updates. The DSL provides constructs for iterative algorithms, property propagation, and graph traversals.

\paragraph{Compiler Infrastructure.}
StarPlat's compiler performs lexical and syntactic analysis, builds an abstract syntax tree (AST), and conducts semantic analysis to ensure type safety and correctness. The intermediate representation captures algorithm semantics independent of target hardware.

\paragraph{Multi-Backend Code Generation.} The StarPlat code generator translates the intermediate representation into parallel implementations targeting graph analytics workloads:
\begin{itemize} 
    \item \textbf{OpenMP}: Generates \texttt{\#pragma omp parallel for} loops to parallelize vertex-centric computations across available CPU cores~\cite{starplat}. 
    \item \textbf{CUDA}: Maps graph topology to the GPU thread hierarchy, assigning vertices or edges to threads to execute purely structural graph traversals~\cite{starplat}. 
    \item \textbf{MPI}: Synthesizes message-passing logic to synchronize scalar vertex data across distributed partitions, utilizing standard graph partitioning libraries~\cite{starplat}. 
    \end{itemize} 
Critically, StarPlat is designed for scalar graph algorithms (e.g., BFS, SSSP) and lacks the specialized memory layouts, SIMD vectorization, and gradient synchronization primitives required for GNN training.

StarPlat achieves competitive performance with hand-optimized implementations on standard graph benchmarks while maintaining code portability. However, it focuses on graph analytics and lacks support for neural network training.

\subsection{Motivation for Morphling}

While StarPlat provides a strong foundation for multi-backend graph processing, GNN workloads introduce new requirements:

\paragraph{Training-Specific Operations.}
GNNs require forward propagation with multiple aggregation schemes (mean, max, sum), automatic differentiation through graph structures, and integration with optimizers (SGD, Adam, AdamW). These operations involve stateful computations, gradient accumulation, and parameter updates absent in traditional graph analytics.

\paragraph{Adaptive Sparsity Handling.}
Graph analytics operate primarily on sparse adjacency matrices with static sparsity patterns. GNN features exhibit dynamic sparsity that varies across datasets, layers, and epochs. Systems must adaptively select dense or sparse execution paths based on runtime profiling to avoid performance degradation.

\paragraph{Limitations of Standard Partitioning}
A critical bottleneck in distributed GNN training is the choice of partitioning strategy. Default implementations in PyG and DGL utilize METIS to minimize edge-cuts. However, as noted in , minimizing edge-cuts is a proxy objective that fails to capture the true cost of GNN execution. In power-law graphs, minimizing edge-cuts can paradoxically maximize the number of unique ghost nodes (remote dependencies) if high-degree hubs are placed on partition boundaries. Furthermore, METIS partitions based on vertex counts, ignoring the fact that computational cost in GNNs is proportional to the number of edges (computational load). This leads to significant workload imbalance, where some ranks idle while others process dense hubs.

\noindent\textbf{Morphling} extends StarPlat to address these challenges. Specifically, it augments the DSL with GNN training primitives and a hierarchical graph partitioner, while implementing a sparsity-aware execution engine that dynamically switches between dense and sparse kernels. Consequently, the system generates optimized training code targeting OpenMP, CUDA, and MPI backends. By leveraging StarPlat's IR as a static foundation while introducing a distributed, hardware-adaptive training runtime, Morphling achieves performance portability for graph learning workloads.


\noindent\section{Morphling DSL}

Morphling bridges the implementation gap across heterogeneous hardware backends by decoupling high-level algorithm specification from low-level execution details. The system operates as a domain-specific code synthesizer that lowers a single declarative program into bespoke implementations for multicore CPUs, many-core GPUs, and distributed clusters. During compilation, Morphling maps abstract GNN constructs to architecture-intrinsic execution models—generating OpenMP dynamic schedules for shared-memory parallelism, CUDA kernels for GPU acceleration, and MPI collectives for distributed synchronization. This abstraction is critical for GNN workloads, where irregular sparsity patterns defy the rigid optimizations of conventional tensor compilers. By automatically instantiating backend-specialized kernels—such as cache-tiled SIMD loops for CPUs and warp-synchronous SpMM for GPUs—Morphling harmonizes fundamentally different parallel paradigms under a unified framework. This ensures consistent scalability and performance portability across diverse memory hierarchies without requiring manual backend tuning.

\begin{lstlisting}[caption={GraphSAGE Training Function in Morphling}, label={lst:list1}]
function SAGE(Graph g, GNN gnn, container<int>& neuronsPerLayer, String Dataset) {
    gnn.load(g, Dataset);
    gnn.initializeLayers(neuronsPerLayer,"xaviers");
    for(int epoch = 0; epoch < totalEpoch; epoch++){
        for(int l = 0; l < gnn.getLayers(); l++)
            gnn.forwardPass(l,"SAGE", "Max");

        for(int l = neuronsPerLayer-1; l >= 0; l--)
            gnn.backPropagation(l);

        gnn.optimizer("adam", 0.01, 0.9, 0.999);
    }    
}
\end{lstlisting}

\subsection{Compiler Overview}

Morphling builds upon the StarPlat~\cite{starplat} compiler frontend, reusing its parser and initial IR construction for graph topology operations. Since StarPlat was designed for static analytics, Morphling extends this infrastructure with \emph{learning-specific state}—gradients, trainable parameters, and optimizer metadata—allowing the compiler to represent full forward–backward GNN pipelines.

\autoref{lst:list1} illustrates the resulting high-level specification. During compilation, these constructs are lowered into a backend-neutral IR, where Morphling schedules the training loop, manages intermediate buffers, and associates each layer with the appropriate compute and communication operators. Rather than emitting low-level code directly, the compiler synthesizes calls into Morphling’s architecture-aware kernel library, enabling backend specialization without exposing complexity to the user.

\begin{itemize}
    \item \texttt{gnn.forwardPass(l, "SAGE", "Max")}: Binds layer~$l$ to the aggregation and transformation operators instantiated for the target backend’s execution model.
    
    \item \texttt{gnn.backPropagation(l)}: Marks the backward boundary for layer~$l$, allowing the compiler to manage gradient buffers more efficiently and keep memory usage under control.
    
    \item \texttt{gnn.optimizer()}: Integrates the chosen update rule (e.g., Adam) into the synthesized training loop, connecting gradient computation with parameter updates without interpreter overhead.
\end{itemize}

This architecture lets the compiler manage control flow and memory lifecycles while delegating computation to highly optimized backend primitives, combining programmability with portable performance across CPU, GPU, and distributed environments.

\begin{figure}[htbp]
    \raggedright
    \includegraphics[width=0.5\textwidth]{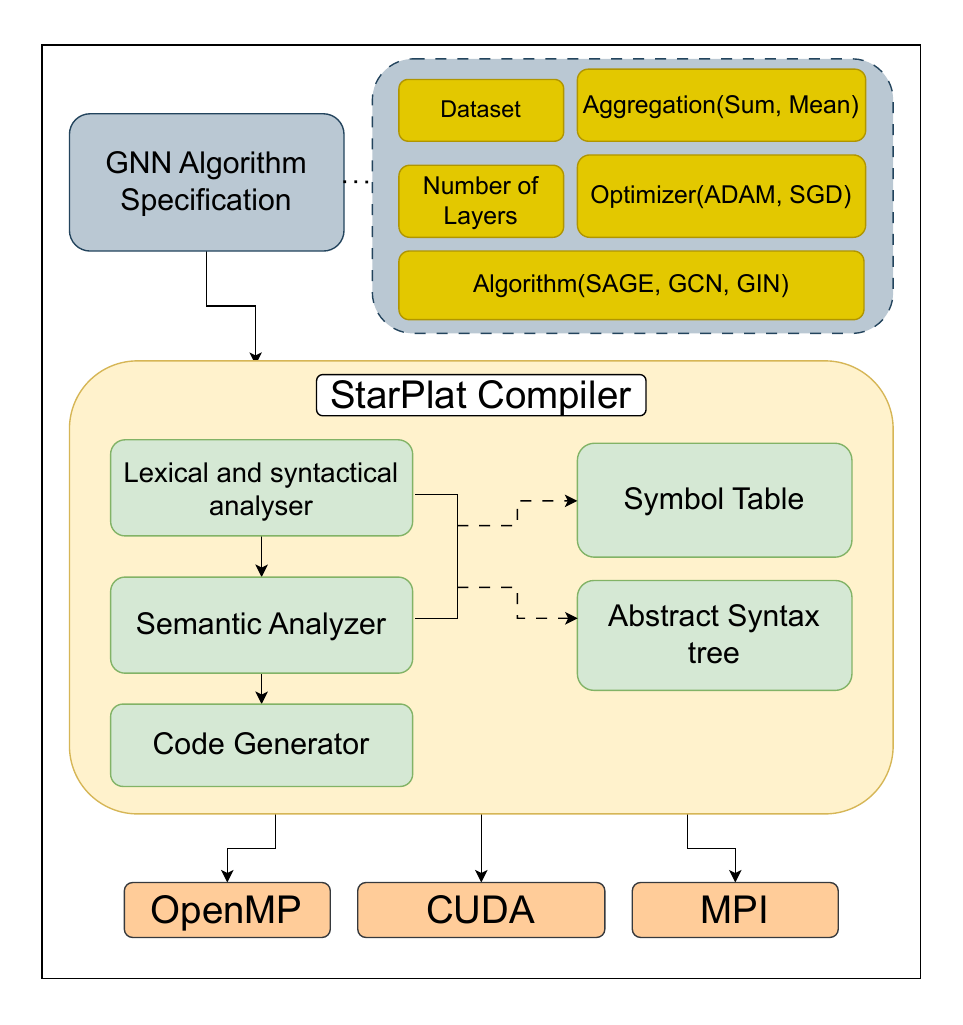}
    \caption{Code Generation framework of Morphling}
    \label{fig:DSL_image}
\end{figure}

\subsection{Sparsity-Aware Execution Engine}
\label{sparse_aware_engine}

Most GNN systems exploit graph sparsity but treat node features as dense tensors. This assumption is suboptimal for real-world workloads where features are intrinsically sparse (e.g., text embeddings). Morphling implements a \emph{dual-path execution engine} that inspects feature statistics at load time and selects the optimal execution strategy—dispatching either to high-throughput dense kernels or memory-efficient sparse primitives.

\paragraph{Sparsity Decision Model.}
Let $X \in \mathbb{R}^{N \times F}$ denote the input feature matrix. The runtime computes the feature sparsity $s = 1 - \frac{\mathrm{nnz}(X)}{N \cdot F}$ during data loading. The decision to switch from dense to sparse execution is guided by the hardware's \textbf{Efficiency Ratio} ($\gamma$).

Let $\eta_{\text{dense}}$ and $\eta_{\text{sparse}}$ represent the measured sustained throughput (FLOPs/s) of the dense GEMM and sparse SpMM kernels, respectively. The sparse path provides a speedup when the reduction in algorithmic work outweighs the lower sustained throughput of sparse kernels (due to irregular memory access). Formally, the sparse path is optimal when:
\begin{equation}
    s > 1 - \gamma, \quad \text{where } \gamma = \frac{\eta_{\text{sparse}}}{\eta_{\text{dense}}}
\end{equation}
On modern CPU architectures and GPUs, dense GEMM is highly optimized, whereas SpMM is bound by indirect memory access latency. Empirical profiling on our testbed reveals that optimized SpMM sustains approximately $20\%$ of the dense baseline throughput ($\gamma \approx 0.20$). This value is obtained from offline microbenchmarks and serves as a default heuristic rather than a per-run calibration. Consequently, Morphling utilizes a tunable threshold $\tau \approx 0.80$. If $s \ge \tau$, the system activates the sparse backend; otherwise, it defaults to the dense path. To our knowledge, existing GNN frameworks do not perform dense–sparse switching for feature matrices.

\paragraph{Static Path Selection}
To minimize runtime overhead, the sparsity check is performed once outside the training loop.
\begin{itemize}
    \item \textbf{Dense Path ($s < \tau$):} The system retains the row-major layout of $X$ and binds the \texttt{forwardPass} directive to vendor-optimized BLAS calls (e.g., \texttt{cblas\_sgemm} or \texttt{cublasSgemm}).
    \item \textbf{Sparse Path ($s \ge \tau$):} The runtime materializes a Compressed Sparse Row (CSR) format for the forward pass and a Compressed Sparse Column (CSC) format for the backward pass. The overhead of this format conversion ($O(\mathrm{nnz})$) is incurred only once during initialization and is amortized over many epochs typical in GNN training.
\end{itemize}

\paragraph{Backend-Specialized Primitives.}
When the sparse path is active, Morphling dispatches to custom kernels designed to maximize bandwidth utilization for irregular accesses:
\begin{itemize}
    \item \textbf{Forward ($X_{csr} \times W$):} On CPUs, we employ a \emph{cache-tiled SpMM} kernel. The weight matrix $W$ is loaded into L1 cache in blocks of size $T_{tile}$, and the sparse feature rows are streamed through these blocks using vectorized fused multiply-add (FMA) instructions. This maximizes temporal locality for $W$ while handling the irregularity of $X$.
    \item \textbf{Backward ($Y \times W^T$):} To support gradient computation, we utilize the pre-computed CSC view of $X$. This allows the backward kernel to iterate over feature columns efficiently, accumulating partial gradients into thread-local buffers before a final reduction, thereby avoiding atomic contention on the shared weight gradient matrix.
\end{itemize}

\paragraph{Crossover Condition 1 (Efficiency Ratio Threshold).}
We define the decision boundary based on the minimization of Time-to-Solution. Let the algorithmic work (FLOPs) for dense and sparse kernels be $W_{\text{dense}} = 2NFH$ and $W_{\text{sparse}} \approx 2(1-s)NFH$, respectively. We model the execution time as $T = W / \eta$, where $\eta$ represents the measured effective throughput (FLOPs/s).

\noindent
\textbf{Derivation.} The sparse path is beneficial when $T_{\text{sparse}} < T_{\text{dense}}$:
\begin{align}
    \frac{W_{\text{sparse}}}{\eta_{\text{sparse}}} &< \frac{W_{\text{dense}}}{\eta_{\text{dense}}} \\
    \frac{2(1-s)NFH}{\eta_{\text{sparse}}} &< \frac{2NFH}{\eta_{\text{dense}}} \\
    \frac{1-s}{\eta_{\text{sparse}}} &< \frac{1}{\eta_{\text{dense}}}
\end{align}
Let $\gamma = \eta_{\text{sparse}} / \eta_{\text{dense}}$ be the \textbf{Efficiency Ratio}, capturing the relative hardware utilization of irregular vs. regular kernels. Rearranging terms yields:
\begin{equation}
    s > 1 - \gamma
\end{equation}
\textit{Interpretation.} The threshold $\tau = 1 - \gamma$ is fully determined by the hardware's ability to handle irregularity. For example, our offline profiling shows that sparse kernels achieve $\approx 20\%$ of dense throughput ($\gamma \approx 0.20$), yielding a theoretical crossover at $s > 0.80$. \textbf{Note:} While the work terms $W$ model algorithmic FLOPs, the efficiency ratio $\gamma$ explicitly absorbs non-algorithmic inefficiencies (e.g., irregular memory access, load imbalance, and cache stalls), making the model robust to micro-architectural variance.

\begin{algorithm}[t]
\small
\caption{Sparsity-Aware Execution Engine}
\label{alg:sparsity_aware_engine}
\begin{algorithmic}[1]
\Require Features $X \in \mathbb{R}^{N \times F}$, Weights $W$, Threshold $\tau \approx 0.80$
\Ensure Output $T$, Gradients $dW$

\Statex \textbf{Phase 1: Runtime Analysis \& Lowering} \Comment{(Once at load)}
\Procedure{Initialize}{$X, \tau$}
    \State $s \gets 1 - \mathrm{nnz}(X)/(N \cdot F)$
    \If{$s \geq \tau$} \Comment{Efficiency check ($\gamma$)}
        \State $\textsc{Mode} \gets \textsc{Sparse}$
        \State $X_{\mathrm{csr}} \gets \textsc{DenseToCSR}(X)$
        \State $X_{\mathrm{csc}} \gets \textsc{DenseToCSC}(X)$ \Comment{For backward pass}
    \Else
        \State $\textsc{Mode} \gets \textsc{Dense}$
    \EndIf
\EndProcedure

\Statex \textbf{Phase 2: Execution} \Comment{(Per-Epoch)}
\Procedure{Forward}{$W$}
    \If{$\textsc{Mode} == \textsc{Sparse}$}
        \State \textbf{return} $\textsc{SpMM\_Tiled}(X_{\mathrm{csr}}, W)$
    \Else
        \State \textbf{return} $\textsc{GEMM}(X, W)$ \Comment{Vendor BLAS}
    \EndIf
\EndProcedure

\Procedure{Backward}{$G$}
    \If{$\textsc{Mode} == \textsc{Sparse}$}
        \State \textbf{return} $\textsc{SpMM\_Col}(X_{\mathrm{csc}}, G)$ \Comment{Conflict-free}
    \Else
        \State \textbf{return} $\textsc{GEMM}^T(X, G)$
    \EndIf
\EndProcedure
\end{algorithmic}
\end{algorithm}

\subsection{CPU Backend: Latency-Hiding Micro-Kernels}

Morphling lowers GNN training to a high-performance CPU backend explicitly targeting the bottlenecks of CSR-based SpMM on modern x86 servers. Two architectural observations motivate our design. First, sparse adjacency induces irregular memory accesses that defeat hardware prefetchers, causing frequent cache and TLB misses. Second, when node feature dimensionality $F$ is large, naive neighbor-wise accumulation traverses full feature vectors for each edge. This leads to poor cache reuse and vector lane under-utilization. To address these issues, Morphling uses a \textit{cache-tiled, software-prefetched SpMM kernel} coupled with dynamic scheduling. The resulting kernels map naturally onto the 64-byte cache lines and 512-bit ZMM registers of the target AVX-512 architecture.

\paragraph{Cache-Aware Feature Tiling.}
The sparse aggregation kernel employs a \emph{feature-tiled} loop nest that operates over the column dimension in contiguous blocks. To eliminate the overhead of runtime loop bound checks, Morphling leverages compile-time template specialization to fix the tile width $T$. In our backend, the kernel is instantiated with $T=32$ (FP32), producing a 128-byte tile—equal in size to two 512-bit AVX-512 vectors—which promotes high SIMD lane utilization. Resolving $T$ at compile time enables the backend compiler to apply constant folding, fully unroll the reduction loop, and generate packed \texttt{vfmadd} micro-ops. The 128-byte tile also aligns cleanly with sequential cache-line access, helping the hardware prefetcher maintain a steady stream of feature data into L1/L2.

For a given target node $u$ and tile offset $k$, the generated kernel:
\begin{enumerate}
    \item \textbf{Register Init:} Zeroes a vector accumulator (ZMM register) for the current tile.
    \item \textbf{Streamed Load:} Iterates over the CSR neighbor list; for each neighbor $v$, loads the contiguous feature block starting at $X[v, k]$.
    \item \textbf{Fused Math:} Performs a vectorized Fused Multiply-Add (FMA) of the neighbor tile, scaled by the edge weight, into the accumulator.
    \item \textbf{Write-Back:} Stores the completed embedding tile to the output $Y[u, k]$.
\end{enumerate}
This layout ensures the inner loop to get packed \texttt{vfmadd} operations, maximizing floating-point pipeline utilization. Tiling increases temporal reuse of neighbor features and aligns memory traffic to cache-line boundaries, which is critical for saturating memory bandwidth.

\paragraph{Software-Pipelined Prefetching.}
To hide DRAM latency induced by irregular neighbor accesses, the kernel issues explicit software prefetch instructions. During neighbor iteration $i$, the kernel issues a \texttt{prefetcht0} hint for the feature address of neighbor $i+D$. Morphling targets the contiguous tile region, enabling the cache hierarchy to bring the relevant 64-byte lines into L1/L2 before consumption. The lookahead distance is tuned (default $D=8$) such that the arithmetic intensity of the intervening FMA operations hides the memory latency. This setup reduces Last-Level Cache (LLC) misses and prevents pipeline stalls. To avoid cache pollution, prefetching is conditionally guarded for low-degree nodes where the setup overhead outweighs the latency benefit.

\begin{algorithm}[t]
\small
\caption{Cache-Tiled SpMM Kernel (CPU)}
\label{alg:csrspmm}
\begin{algorithmic}[1]
\Require Graph $G$ (CSR), Features $X$, Tile Width $T$
\Ensure Output Embedding $Y$

\Statex \textbf{Phase 1: Initialization}
\State $Y[0 \dots N \cdot F] \leftarrow 0$ \Comment{Parallel Bulk Zero via SIMD}

\Statex \textbf{Phase 2: Tiled Aggregation}
\For{each node $u \in 0 \dots N-1$ \textbf{in parallel}} \Comment{OpenMP Dynamic Schedule}
    \For{$ei = \mathrm{row\_ptr}[u]$ \textbf{to} $\mathrm{row\_ptr}[u+1]$} \Comment{Neighbor Loop}
        \State $v \gets \mathrm{col\_idx}[ei]$; \quad $w \gets \mathrm{val}[ei]$

        \If{$ei + D < \mathrm{row\_ptr}[u+1]$} \Comment{Software Prefetching}
            \State $v_{next} \gets \mathrm{col\_idx}[ei+D]$
            \State $\textsc{PrefetchL1}(X[v_{next}, :])$
        \EndIf

        \For{$f = 0$ \textbf{to} $F$ \textbf{step} $T$} \Comment{Feature Tile Loop}
            \State $\mathbf{y}_{\text{reg}} \gets \textsc{LoadZMM}(Y[u, f:f+T])$
            \State $\mathbf{x}_{\text{reg}} \gets \textsc{LoadZMM}(X[v, f:f+T])$
            \State $\mathbf{y}_{\text{reg}} \gets \textsc{VFMADD}(\mathbf{y}_{\text{reg}}, \mathbf{x}_{\text{reg}}, w)$ \Comment{Fused Multiply-Add}
            \State $\textsc{StoreZMM}(Y[u, f:f+T], \mathbf{y}_{\text{reg}})$
        \EndFor
    \EndFor
\EndFor
\end{algorithmic}
\end{algorithm}

\paragraph{Dynamic Scheduling.}
To mitigate load imbalance from power-law degree distributions, Morphling uses the outer node loop to an OpenMP parallel region with a dynamic schedule (e.g., \texttt{schedule(dynamic,64)}). This strategy hands out fine-grained work chunks to threads at runtime, ensuring that heavily connected hubs do not stall the worker threads. The dense GEMM path is delegated to vendor-optimized BLAS (e.g., Intel MKL), while the sparse path parallelizes across nodes (row-wise) for forward aggregation and across feature columns for gradient computation, structuring the execution to avoid atomic write conflicts by design.

\subsection{CUDA Backend} 
\label{sec:cuda_backend} 

The CUDA backend employs a throughput-oriented execution model that aims to achieve high effective memory bandwidth utilization. Morphling dispatches operations to a library of custom CUDA kernels that map graph primitives directly onto the GPU thread hierarchy, prioritizing memory coalescing and minimal intermediate allocation. While the Block-per-Row (BPR) mapping is a standard strategy for sparse GPU kernels, Morphling integrates it into a unified backend design so that the same feature-parallel decomposition used on CPUs carries over to GPUs, ensuring consistent execution semantics across architectures. The key contribution of Morphling’s CUDA backend is that it unifies dense–sparse execution and feature-parallel decomposition across CPU and GPU targets, allowing the compiler to select kernels automatically based on runtime sparsity without changing the high-level DSL.

\paragraph{Block-Parallel Row Mapping.}
For the core SpMM aggregation, Morphling adopts the \textbf{Block-per-Row} execution strategy wherein each output node $u$ is mapped to a single CUDA thread block (\texttt{blockIdx.x}). Within the block, threads parallelize the computation along the feature dimension $F$ using a block-strided loop. This mapping secures two key architectural properties:

\begin{enumerate}
    \item \textbf{Reduction of Control Divergence:} Since the neighbor range ($\text{row\_ptr}[u]$ to $\text{row\_ptr}[u+1]$) is identical for the entire block, all threads within a warp execute the neighbor traversal loop in lockstep. This structural uniformity removes the degree-induced divergence penalties that affect thread-per-node mappings.
    \item \textbf{Atomic-Free Accumulation:} Each thread computes a disjoint set of feature indices $Y[u, f + k \cdot \mathrm{stride}]$, accumulating partial results in thread-local registers before a single write-back to global memory. By ensuring no two threads update the same output element, the kernel avoids the high-latency atomic operations required by scatter-gather designs.
\end{enumerate}

Although this mapping can expose inter-block load imbalance due to power-law degree distributions, real-world GNN workloads offer thousands of independent rows per launch. This allows the SM scheduler to effectively distribute blocks, overlapping short-running leaf blocks with long-running hub blocks to maintain device occupancy.

\paragraph{Implicit-Transpose Gradient (Backward Pass).}
Standard frameworks typically materialize a Transposed (CSC) copy of the sparse feature matrix to compute weight gradients ($dW = X^T G$). While effective for compute throughput, this doubles the memory footprint of the graph structure. Morphling adopts a memory-centric engineering trade-off: instead of materializing the CSC, we employ an Implicit-Transpose Kernel.
The backend iterates over the original CSR structure. For each non-zero entry, threads compute partial gradients and scatter them to the weight gradient matrix using \texttt{atomicAdd}. While this introduces atomic contention, it allows gradients to be computed in-place. This technique, combined with the fusion of message passing (avoiding $O(|E|)$ intermediate tensors), drives the significant peak memory reduction observed in our evaluation, enabling the training of massive graphs on commodity GPUs where capacity is the primary bottleneck.

\begin{algorithm}[t]
\small
\caption{Atomic-Free SpMM Aggregations}
\label{alg:atomic_free_spmm}
\begin{algorithmic}[1]
\Require Graph $G$ (CSR), Input $X$, Feature Dim $F$
\Ensure Output $Y$ \Comment{Computed without atomics}

\Statex \textbf{Grid-Level Parallelism}
\For{each block $u \in 0 \dots N-1$ \textbf{in parallel}} \Comment{1 Block per Node}
    \State $\mathrm{start} \gets \mathrm{row\_ptr}[u]$; \quad $\mathrm{end} \gets \mathrm{row\_ptr}[u+1]$

    \Statex \textbf{Thread-Level Parallelism}
    \For{$f = \mathrm{tid}$ \textbf{to} $F$ \textbf{step} $\mathrm{blockDim}$} \Comment{Grid-Stride Loop}
        \State $\mathrm{acc} \gets 0$ \Comment{Register Accumulator}

        \For{$ei = \mathrm{start}$ \textbf{to} $\mathrm{end}$} \Comment{Lockstep Neighbor Scan}
            \State $v \gets \mathrm{col\_idx}[ei]$; \quad $w \gets \mathrm{val}[ei]$
            \State $\mathrm{acc} \gets \mathrm{acc} + w \cdot X[v, f]$ \Comment{Coalesced Read}
        \EndFor

        \State $Y[u, f] \gets \mathrm{acc}$ \Comment{Conflict-Free Write}
    \EndFor
\EndFor
\end{algorithmic}
\end{algorithm}


\subsection{MPI Backend}
Real-world graph datasets frequently exceed the memory capacity of a single node, necessitating distributed GNN training. Morphling addresses this requirement through an MPI-based backend that partitions both computation and data across machines while preserving the DSL’s high-level programming model. The runtime applies a hierarchical partitioner to construct balanced subgraphs, materializes local and ghost-node views, and generates the communication schedules required for aggregation and gradient synchronization. During training, Morphling overlaps inter-node communication with computation using non-blocking collectives and point-to-point exchanges, enabling scalable execution without exposing low-level MPI details to the user.



\subsubsection{Data Partitioner}
\label{sec:partitioner}

The performance of distributed GNN training is governed by two conflicting objectives: minimizing the \textit{edge-cut ratio} (to reduce communication volume) and minimizing \textit{computational imbalance} (to prevent straggler effects). Standard partitioning libraries, such as METIS, typically optimize for vertex-count balance. This approach often fails to account for the power-law degree distributions inherent in real-world graphs, where a partition with equal vertex counts can still suffer from severe computational imbalance if one rank processes a disproportionate number of high-degree hubs.

To address this, Morphling implements an Adaptive Hierarchical Partitioning engine (Algorithm~\ref{alg:partition}) that treats partitioning as a constraint satisfaction problem with progressively relaxing constraints. The engine operates in three distinct phases:

\begin{enumerate}[leftmargin=*]
    \item \textbf{Phase I: Topology-Aware Minimization.} 
    The system first attempts to solve the $k$-way partitioning problem using METIS to minimize edge-cuts subject to a strict load imbalance constraint $\epsilon=1.03$. We prioritize the SHEM (Sorted Heavy Edge Matching) coarsening strategy to maximize local neighborhood preservation. If convergence fails---often due to disconnected components or small-world structures---the system automatically relaxes the imbalance tolerance ($\epsilon \to 1.20$) and switches to recursive bisection, which exhibits higher stability on smaller, irregular graphs.

    \item \textbf{Phase II: Component-Aware Bin Packing.} 
    If topological partitioning fails to converge, the system detects disconnected subgraphs (connected components) via BFS traversal. It then employs a \textbf{Best-Fit Decreasing} bin-packing heuristic: components are sorted by size ($|V_C|$) and assigned to partitions to minimize the variance in partition sizes: 
    \begin{equation}
        \min \sum_{p} |V_p - \bar{V}|^2
    \end{equation}
    This phase ensures that naturally dense subgraphs remain local to a single rank, reducing cross-partition communication for those regions.

    \begin{table}[t]
    \centering
    \caption{Comparison of Partitioning Strategies}
    \label{tab:complexity}
    \small
    \newcolumntype{L}{>{\raggedright\arraybackslash}X}
    
    \begin{tabularx}{\columnwidth}{@{}l L l L@{}}
    \toprule
    \textbf{Strategy} & \textbf{Primary Objective} & \textbf{Time} & \textbf{Dominant Cost} \\
    \midrule
    METIS & Edge-Cut Min. & $O(|E|)$ & Multilevel Coarsening \\
    \addlinespace
    Component & Connectivity & $O(|V|{+}|E|)$ & Graph Traversal \\
    \addlinespace
    Greedy & Load Balance & $O(|V| \log |V|)$ & Degree Sorting \\
    \bottomrule
    \multicolumn{4}{@{}p{\columnwidth}@{}}{\scriptsize{\textit{Note:} $|V|$: Nodes, $|E|$: Edges. METIS time varies by configuration.}}
    \end{tabularx}
    \end{table}

    \item \textbf{Phase III: Load-Aware Greedy Fallback.} 
    For pathological graphs where topological cuts are infeasible (e.g., massive star graphs with high-degree hubs), Morphling falls back to a greedy degree-based strategy. Vertices are sorted by degree ($d_v$) in descending order and assigned to the partition with the minimum current weight. Crucially, unlike standard approaches that balance based on vertex count ($weight_p = |V_p|$), this phase explicitly balances computational load:
    \begin{equation}
        weight_p = \sum_{v \in P} d_v
    \end{equation}
    By distributing high-degree hubs first, this strategy prevents the formation of ``straggler partitions'' that would otherwise bottleneck the synchronous distributed barrier.
\end{enumerate}

\begin{algorithm}[t]
\small 
\caption{Hierarchical Constraint Relaxation Partitioning}
\label{alg:partition}
\begin{algorithmic}[1]
\Require Graph $G(V, E)$, Partitions $k$
\Ensure Partition Map $P: V \to \{0, \dots, k-1\}$

\State \textbf{Phase I: Topology-Aware Minimization}
\State $opts \leftarrow$ \textsc{InitMetis}(\texttt{SHEM}, $\epsilon=1.03$)
\State $status \leftarrow$ \textsc{MetisKWay}($G, k, opts$)
\If{$status \neq$ OK}
    \State $opts.\epsilon \leftarrow 1.20$ \Comment{Relax Imbalance Constraint}
    \State $status \leftarrow$ \textsc{MetisRecursive}($G, k, opts$)
\EndIf

\If{$status =$ OK}
    \State \Return $P$
\EndIf

\State \textbf{Phase II: Component Bin Packing}
\State $Comps \leftarrow$ \textsc{FindComponents}($G$)
\If{$|Comps| > 1$}
    \State Sort $Comps$ by size descending
    \State $Weights[0 \dots k] \leftarrow 0$
    \For{$C \in Comps$}
        \State $p \leftarrow \arg\min(Weights)$ \Comment{Best-Fit Assignment}
        \State $P[v] \leftarrow p, \forall v \in C$
        \State $Weights[p] \leftarrow Weights[p] + |C|$
    \EndFor
    \State \Return $P$
\EndIf

\State \textbf{Phase III: Load-Aware Greedy Fallback}
\State \textit{// Optimizes $\sum deg(v)$ instead of $|V|$}
\State $Nodes \leftarrow$ Sort $V$ by Degree descending
\State $Weights[0 \dots k] \leftarrow 0$
\For{$v \in Nodes$}
    \State $p \leftarrow \arg\min(Weights)$ 
    \State $P[v] \leftarrow p$
    \State $Weights[p] \leftarrow Weights[p] + \text{deg}(v) + 1$
\EndFor
\State \Return $P$
\end{algorithmic}
\end{algorithm}

\subsubsection{Distributed Runtime Implementation}
\label{sec:mpi_runtime}

The Morphling MPI backend is architected as a Bulk Synchronous Parallel (BSP) system with explicit communication-computation overlap. To mitigate the high latency of distributed interconnects, the runtime employs a split-phase communication model for both feature synchronization and gradient aggregation.

\begin{enumerate}[leftmargin=*]
    \item \textbf{Local-Global Address Translation.} 
    To support efficient sparse operations without expensive hash lookups during the training loop, the runtime materializes a thread-safe \textit{Global-to-Local (G2L)} mapping. Each rank maintains a contiguous memory space where local nodes $[0, N_{local})$ are followed immediately by ghost nodes. This layout enables the SpMM kernels to operate on dense, contiguous index ranges, permitting the usage of AVX-512 vectorization on the local feature tensors.

    \item \textbf{Asynchronous Halo Exchange.} 
    Feature synchronization relies on non-blocking point-to-point primitives. As seen in the \texttt{exchange\_ghost} routine, we employ a multi-stage protocol:
    \begin{itemize}
        \item \textit{Parallel Packing:} OpenMP threads gather non-contiguous features from the embedding table into pre-allocated send buffers.
        \item \textit{Non-Blocking Issue:} Requests are dispatched via \texttt{MPI\_Isend} / \texttt{MPI\_Irecv}, allowing the network interface controller (NIC) to handle data transfer while the CPU performs initialization for the forward pass.
        \item \textit{Wait-Free Unpacking:} Received data is unpacked directly into the ghost region of the feature matrix, making it immediately available for aggregation.
    \end{itemize}

    \item \textbf{Pipelined Backward Propagation.} 
    A critical optimization in Morphling is the masking of global synchronization latency. Standard distributed implementations often block immediately after computing weight gradients. In contrast, Morphling utilizes a \textit{Gradient Communication Pipeline} (implemented in \texttt{backward\_layer\_pipelined}):
    \begin{enumerate}
        \item \textbf{Compute $dW$:} The kernel calculates local weight gradients $dW_{local}^{(l)}$ using the transposed input features.
        \item \textbf{Issue Reduction:} The runtime immediately issues a non-blocking \texttt{MPI\_Iallreduce}.
        \item \textbf{Overlap ($dX$):} While the reduction traverses the network fabric, the CPU concurrently computes the gradients with respect to the input activations ($dX^{(l-1)}$). This operation typically dominates the layer execution time, effectively hiding the cost of parameter synchronization.
        \item \textbf{Synchronize:} The runtime calls \texttt{MPI\_Wait} only after the input gradient computation completes, ensuring the optimizer receives the fully aggregated global gradients.
    \end{enumerate}

    \item \textbf{Vectorized Optimizer.} 
    Parameter updates are executed via a custom AVX-optimized kernel (\texttt{adam\_update\_vectorized}). Unlike frameworks that rely on Python-level loops or external tensor libraries for optimization, Morphling keeps model weights in C++ memory. It applies fused momentum and variance updates via SIMD pragmas immediately after the synchronization barrier, minimizing memory traffic and preventing cache thrashing.
\end{enumerate}

\subsubsection{Communication Complexity Analysis}
\label{sec:complexity}

To rigorize the necessity of the multi-objective partitioning strategy described in Section~\ref{sec:partitioner}, we formulate the theoretical execution cost of the distributed training loop derived from our MPI implementation.

Let $G=(V, E)$ be partitioned across $P$ ranks. The wall-clock time for a single training epoch, $T_{epoch}$, is bounded by the slowest rank (the straggler):
\begin{equation}
    T_{epoch} = \max_{p \in \{0 \dots P-1\}} \left( T_{comp}^{(p)} + T_{comm\_halo}^{(p)} + T_{comm\_grad}^{(p)} \right)
\end{equation}

\noindent\textbf{1) Computational Cost ($T_{comp}^{(p)}$):}
The dominant operation is the Sparse Matrix-Matrix multiplication (SpMM) in the aggregation phase. As shown in our kernel implementation, this scales linearly with the number of local edges:
\begin{equation}
    T_{comp}^{(p)} \propto \sum_{v \in V_{local}^{(p)}} \text{deg}(v) \times F
\end{equation}
where $F$ is the feature dimensionality. This explicitly justifies \textbf{Phase III (Greedy Fallback)} of our partitioner. Standard partitioning balances $|V_{local}|$, but Morphling optimizes $\sum \text{deg}(v)$, ensuring that $T_{comp}^{(p)}$ is uniform across ranks.

\noindent\textbf{2) Halo Exchange Cost ($T_{comm\_halo}^{(p)}$):}
Feature synchronization requires point-to-point transfers of ghost node embeddings. Utilizing the LogP model, the cost is:
\begin{equation}
    T_{comm\_halo}^{(p)} \approx \sum_{r \neq p} \left( \alpha + \beta \cdot |V_{halo}^{(p,r)}| \cdot F \right)
\end{equation}
where $|V_{halo}^{(p,r)}|$ is the number of boundary nodes shared between rank $p$ and $r$. This term is directly proportional to the \textbf{Edge-Cut} metric. This justifies \textbf{Phase I (METIS)} of our partitioner, which explicitly minimizes $\sum |E_{cut}|$ to reduce bandwidth pressure ($\beta$ term).

\noindent\textbf{3) Gradient Aggregation ($T_{comm\_grad}^{(p)}$):}
Weight gradients are synchronized via \texttt{MPI\_Allreduce}. Assuming a Ring-AllReduce algorithm:
\begin{equation}
    T_{comm\_grad}^{(p)} \approx 2(P-1)\alpha + 2\frac{P-1}{P} \beta |W|
\end{equation}
Since the model size $|W|$ is constant and typically small for GNNs, this term is negligible compared to halo exchange for large graphs, allowing our system to focus optimization efforts on the data-dependent terms ($T_{comp}$ and $T_{halo}$).

\section{Experimental Evaluation}
\label{sec:evaluation}

\subsection{Datasets}
We evaluate Morphling on ten real-world graph datasets detailed in Table~\ref{tab:datasets}. These benchmarks were selected to span a broad spectrum of topological characteristics, ranging from small, dense citation networks (e.g., \textit{Corafull}, \textit{Physics}) to large-scale, sparse social and e-commerce graphs (e.g., \textit{Reddit}, \textit{AmazonProducts}).

The datasets exhibit diverse properties that stress different components of the training pipeline:
\begin{itemize}
    \item \textbf{Scale Variance:} Graph sizes range from 19K to 2.4M nodes and up to 264M edges, allowing us to evaluate throughput scaling from cache-resident workloads to memory-bound scenarios.
    \item \textbf{Feature Dimensionality:} Feature widths range from 50 to 61,278 (NELL), testing the system's ability to handle both topology-bound and feature-bound computation.
    \item \textbf{Sparsity Regimes:} We include graphs with varying degrees of feature sparsity and edge density to validate the effectiveness of our sparsity-aware execution engine.
\end{itemize}

All experiments were conducted on the full datasets. Notably, the \textit{AmazonProducts} graph (264M edges) exceeds the memory capacity of standard PyG implementations on our testbed. Consequently, we report PyG as Out-of-Memory (OOM) for this benchmark, highlighting Morphling's superior memory efficiency.

\begin{table}[t]
\centering
\caption{Dataset Statistics}
\label{tab:datasets}
\small
\begin{tabular}{@{}lrrrr@{}}
\toprule
\textbf{Dataset} & \textbf{Nodes} & \textbf{Edges} & \textbf{Features} & \textbf{Classes} \\
\midrule
Corafull & 19,793 & 126,842 & 8,710 & 70 \\
Physics & 34,493 & 495,924 & 8,415 & 5 \\
PPI & 56,944 & 1,612,348 & 50 & 121 \\
NELL & 65,755 & 251,550 & 61,278 & 186 \\
Flickr & 88,250 & 899,756 & 500 & 7 \\
Reddit & 232,965 & 114,615,892 & 602 & 41 \\
Yelp & 716,847 & 13,954,819 & 300 & 100 \\
AmazonProducts & 1,569,960 & 264,339,468 & 200 & 107 \\
ogbn-arxiv & 169,343 & 1,166,243 & 128 & 40 \\
ogbn-products & 2,449,029 & 61,859,140 & 100 & 47 \\
\bottomrule
\end{tabular}
\end{table}

\subsection{Hardware and Software Configuration}
\label{sec:hardware}
All experiments were conducted on the \textit{PARAM Rudra} supercomputer at the High Performance Computing and Data Centre (HPCE), IIT Madras. The cluster employs a Lustre parallel file system for high-throughput data access and uses SLURM for resource management. The specific configurations are as follows:

\begin{itemize}
    \item \textbf{CPU Testbed:} Compute nodes are equipped with dual-socket Intel Xeon Gold 6240R (Cascade Lake) processors (48 cores per node) and 192~GB of DDR4 system memory.
    \item \textbf{GPU Testbed:} Accelerated nodes contain dual NVIDIA A100-PCIE GPUs (40~GB HBM2e) and dual Intel Xeon Gold 6240R processors with 192~GB of DDR4 RAM. To mitigate I/O bottlenecks, each node is equipped with an 800~GB local NVMe SSD for caching graph partitions.
    \item \textbf{Distributed Interconnect:} Nodes are connected via a low-latency, high-bandwidth InfiniBand NDR fabric, which is utilized by our MPI backend for gradient and feature synchronization.
    \item \textbf{Software Environment:} Nodes run Linux (Alma 8.9). Morphling is compiled with GCC 9.4 and CUDA 11.8. We compare against \textbf{PyTorch Geometric (v2.6.1)} and \textbf{DGL (v2.1.0)} using PyTorch 2.0.0 as the backend.
    \item \textbf{Model Architecture:} For all benchmarks, we utilize a 3-layer Graph Convolutional Network (GCN) with a hidden dimension of 32.
\end{itemize}

\begin{table}[t]
    \centering
    \caption{Peak System Memory Consumption (GB). Morphling maintains a compact footprint, whereas PyG exhibits excessive memory expansion due to intermediate tensor materialization.}
    \label{tab:memoryusage}
    \small
    \begin{tabular}{@{}lrrr@{}}
        \toprule
        \textbf{Dataset} & \textbf{Morphling} & \textbf{PyG} & \textbf{DGL} \\
        \midrule
        Reddit         & \textbf{4.40} & 48.12 & 8.75 \\
        Yelp           & \textbf{2.59} & 14.01 & 5.04 \\
        AmazonProducts & \textbf{9.03} & 140.24\textsuperscript{\textdagger} & 25.10 \\
        ogbn-arxiv     & 0.57 & 1.14 & \textbf{0.34} \\
        ogbn-products  & \textbf{7.00} & 57.12 & 11.84 \\
        \bottomrule
        \multicolumn{4}{@{}p{\columnwidth}@{}}{\footnotesize{\textsuperscript{\textdagger}\,\textbf{Failure Case:} PyG failed to process the full AmazonProducts graph due to memory exhaustion (OOM $>192$ GB). The reported 140.24 GB is a lower-bound measurement taken on a 75\% subsampled graph.}}
    \end{tabular}
\end{table}

\subsection{Morphling on CPU}
\label{sec:cpu_eval}

\begin{figure}[htbp]
    \raggedright
    \includegraphics[width=0.5\textwidth]{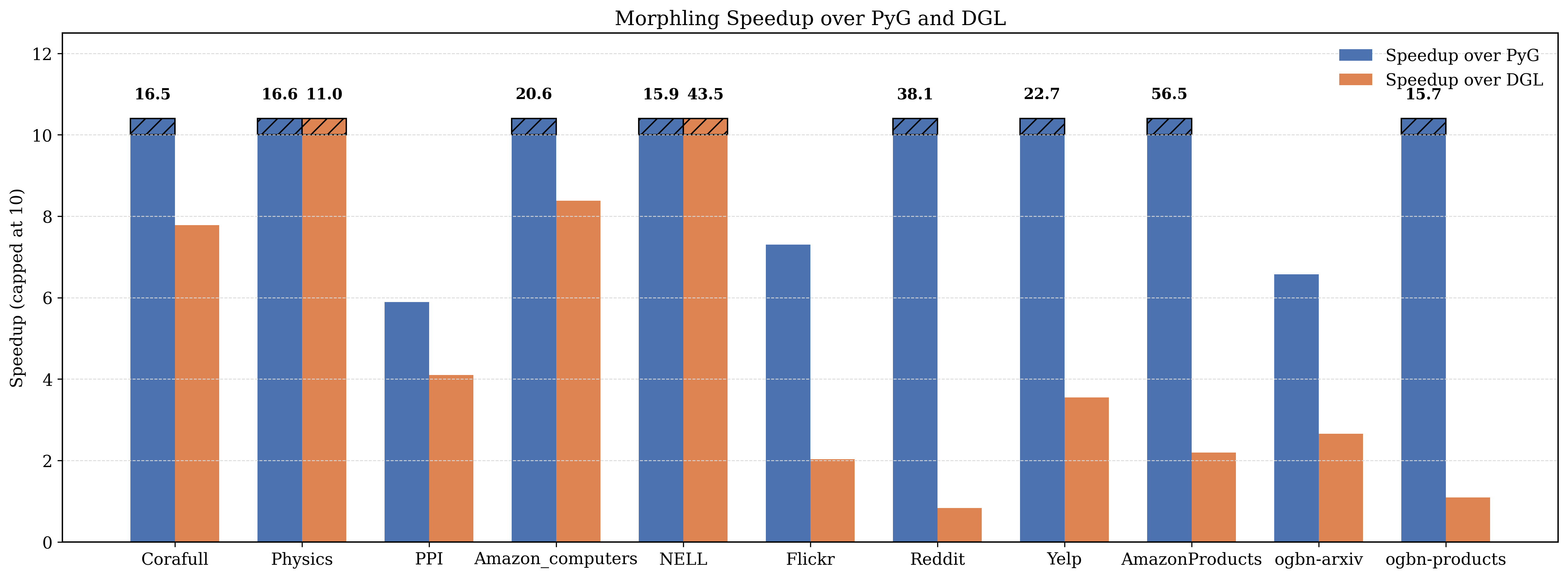}
    \caption{Morphling speedup over PyG and DGL on CPU}
    \label{fig:cpuspeedup}
\end{figure}

\begin{figure}[htbp]
    \raggedright
    \includegraphics[width=0.5\textwidth]{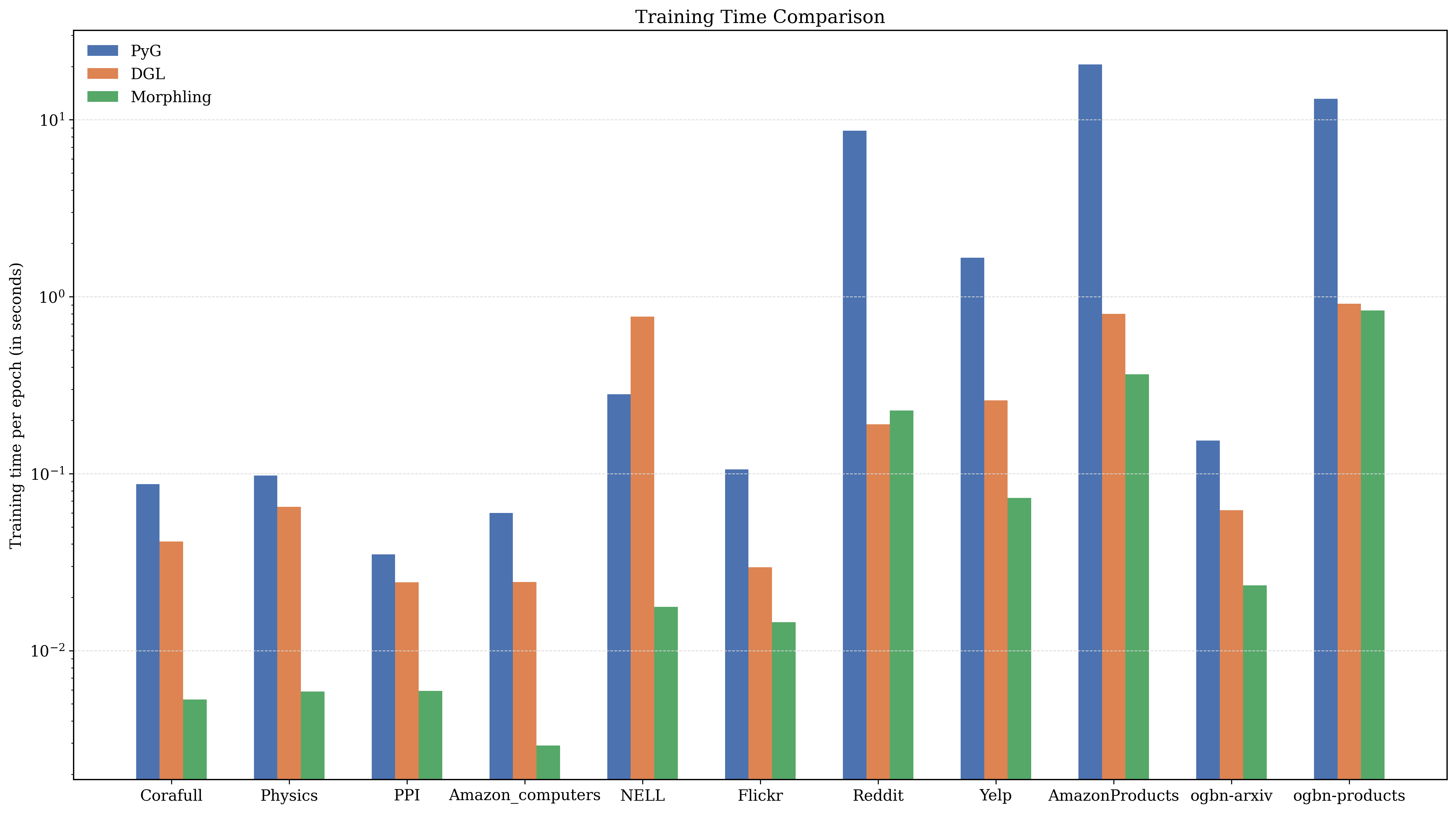}
    \caption{Per-epoch training time of Morphling, PyG, and DGL on CPU}
    \label{fig:cputraining}
\end{figure}

We evaluate Morphling under full-batch CPU training and compare its sustained throughput against PyG and DGL. Figure~\ref{fig:cpuspeedup} reports relative speedups, while Figure~\ref{fig:cputraining} presents absolute per-epoch latencies.

\subsubsection{Metric Definition}
We measure \textbf{sustained per-epoch training time}, which includes the forward pass, backward pass, and optimizer update. Morphling’s one-time code generation and PyTorch/DGL JIT warm-up are excluded, as these costs are amortized over long training runs and do not affect steady-state throughput.

\subsubsection{Performance Analysis}
Morphling achieves the highest throughput on 9 of the 10 datasets. On average, it delivers a \textbf{20.21$\times$} speedup over PyG and \textbf{8.20$\times$} over DGL. The only exception is \textit{Reddit}, where dense feature vectors favor DGL’s MKL-backed dense kernels; here Morphling achieves a competitive \textbf{0.23 s/epoch} versus DGL’s \textbf{0.19 s/epoch}.

On sparse or high-dimensional graphs, however, Morphling holds a decisive advantage. For instance, \textit{NELL} exhibits \textbf{99.21\% feature sparsity}, incur redundant dense operations on mostly zero-valued entries. Morphling’s sparsity-aware dispatch reduces epoch latency by \textbf{43.52$\times$} relative to DGL. Similarly, on \textit{AmazonProducts}, Morphling completes an epoch in \textbf{0.36 s}, whereas PyG fails to execute due to memory exhaustion (OOM) and DGL requires \textbf{0.80 s}.

These results validate the design principles behind Morphling:
\begin{enumerate}
    \item \textbf{Sparse vs.\ Dense Regimes.}  
    Morphling detects feature sparsity (experimentally tuned threshold $s \approx 0.85$) and routes computation to a specialized tiled SpMM kernel, avoiding the redundant FLOPs incurred by dense GEMMs.
    
    \item \textbf{Kernel Specialization.}  
    Morphling’s AVX-512–optimized kernels maintain contiguous memory access and avoid the pointer-chasing overhead of generic sparse-tensor routines (e.g., PyG’s scatter/gather). This reduces per-epoch latency from seconds (8.69 s on Reddit for PyG) to sub-second performance.
\end{enumerate}

In summary, while dense optimization gives DGL a slight edge on the Reddit graph, Morphling consistently provides superior throughput across the broader spectrum of real-world sparse and irregular graphs.

\subsection{Morphling on GPU}
\label{sec:gpu_eval}

\begin{figure}[htbp]
    \raggedright
    \includegraphics[width=0.5\textwidth]{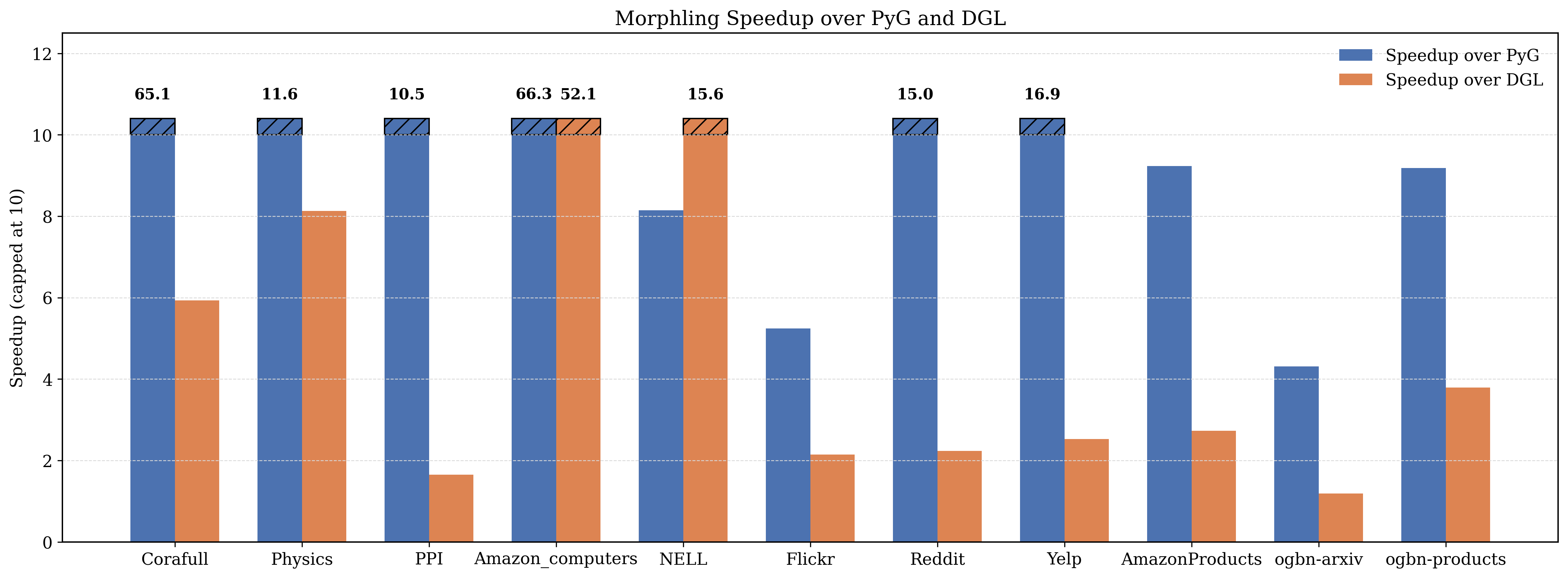}
    \caption{Morphling speedup over PyG and DGL on GPU}
    \label{fig:gpuspeedup}
\end{figure}

\begin{figure}[htbp]
    \raggedright
    \includegraphics[width=0.5\textwidth]{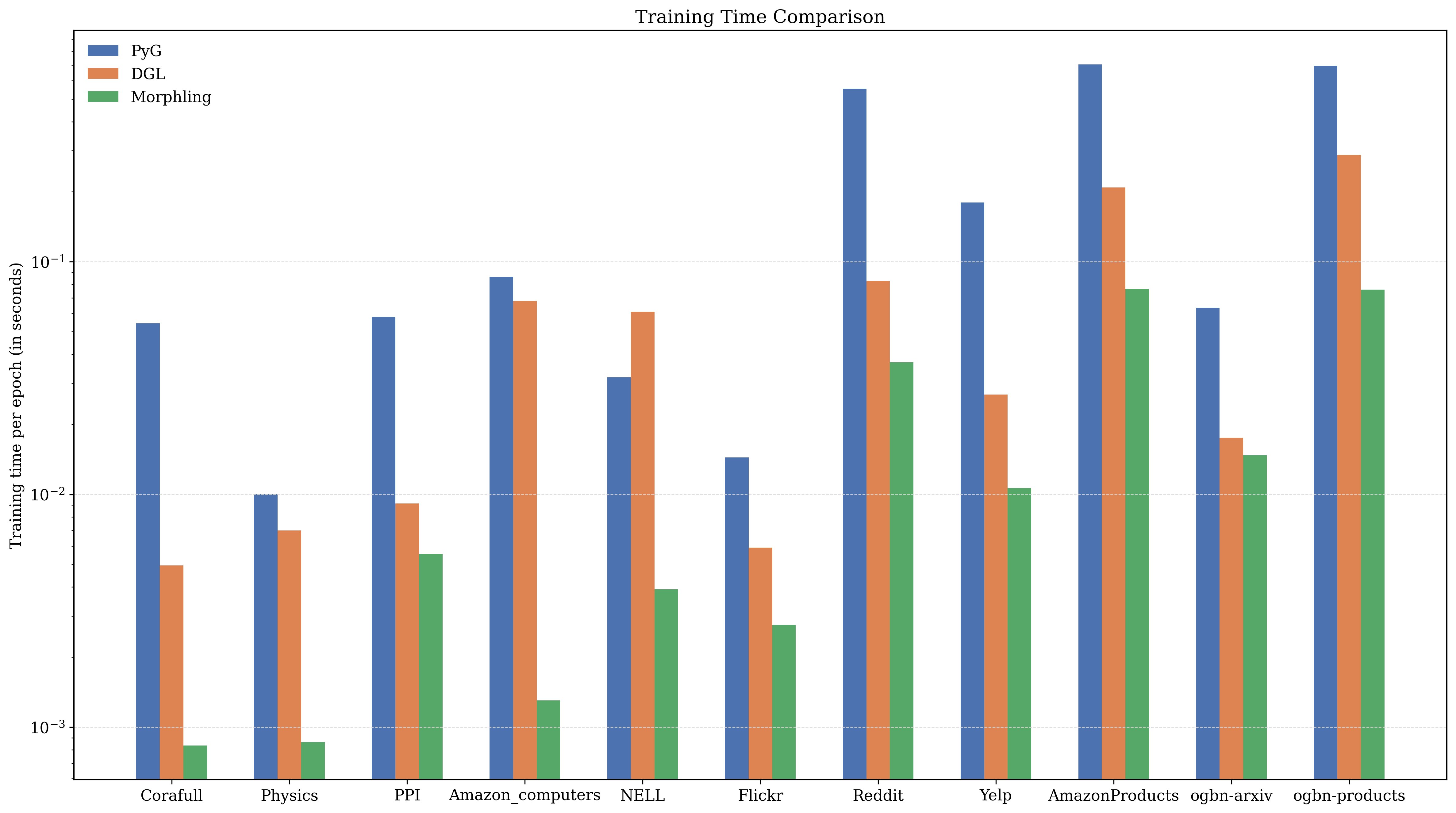}
    \caption{Per-epoch training time of Morphling, PyG, and DGL on GPU}
    \label{fig:gputraining}
\end{figure}

We evaluate Morphling on the NVIDIA A100 under full-batch training. 
Figure~\ref{fig:gpuspeedup} reports relative speedups, while 
Figure~\ref{fig:gputraining} shows per-epoch latencies.

\subsubsection{Throughput Analysis}
Morphling provides substantial GPU throughput gains across all datasets, achieving an average \textbf{15.48$\times$} speedup over PyG and \textbf{4.40$\times$} over DGL. The largest improvements arise on sparse or irregular graphs. On \textit{AmazonComputers}, Morphling reduces epoch time from \textbf{17.26 s} (PyG) and \textbf{13.56 s} (DGL) to \textbf{0.26 s}, corresponding to \textbf{66.28$\times$} and \textbf{52.06$\times$} speedups.

A critical case is \textit{AmazonProducts}. PyG fails on the full dataset due to GPU memory exhaustion, so we report PyG on a \textbf{55\% subsampled graph}. Even under this favorable condition, PyG requires \textbf{140.93 s/epoch}, while DGL processes the full graph in \textbf{41.73 s}. Morphling completes the full dataset in \textbf{15.26 s}, providing a \textbf{2.73$\times$} speedup over DGL and an effective \textbf{9.23$\times$} speedup over PyG despite the subsampling advantage.

On bandwidth-intensive graphs such as \textit{Reddit}, Morphling continues to outperform both baselines, achieving a \textbf{2.23$\times$} speedup over DGL and \textbf{14.99$\times$} over PyG.

\subsubsection{Architectural Attribution}
Morphling’s GPU gains stem from two factors:  
(1) a \textit{Block-Per-Row} kernel that ensures warp-coalesced feature access and avoids PyG’s irregular scatter/gather patterns, and  
(2) fused aggregation kernels that eliminate edge-proportional message tensors created by PyG and DGL. Together, these optimizations reduce global memory traffic and enable execution on large graphs where baselines either stall or OOM.

\subsection{Morphling on MPI}
\label{sec:mpi_eval}

\begin{figure}[htbp]
    \raggedright
    \includegraphics[width=0.5\textwidth]{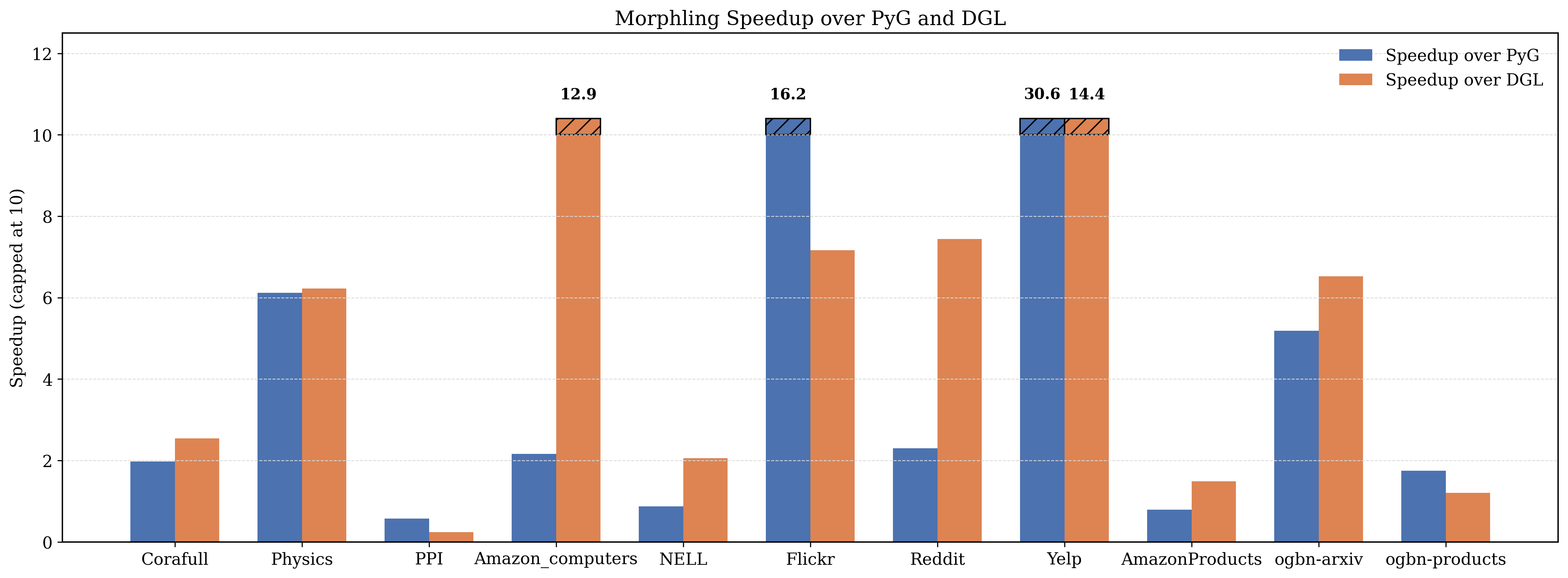}
    \caption{Morphling speedup over PyG and DGL on MPI}
    \label{fig:mpispeedup}
\end{figure}

\begin{figure}[htbp]
    \raggedright
    \includegraphics[width=0.5\textwidth]{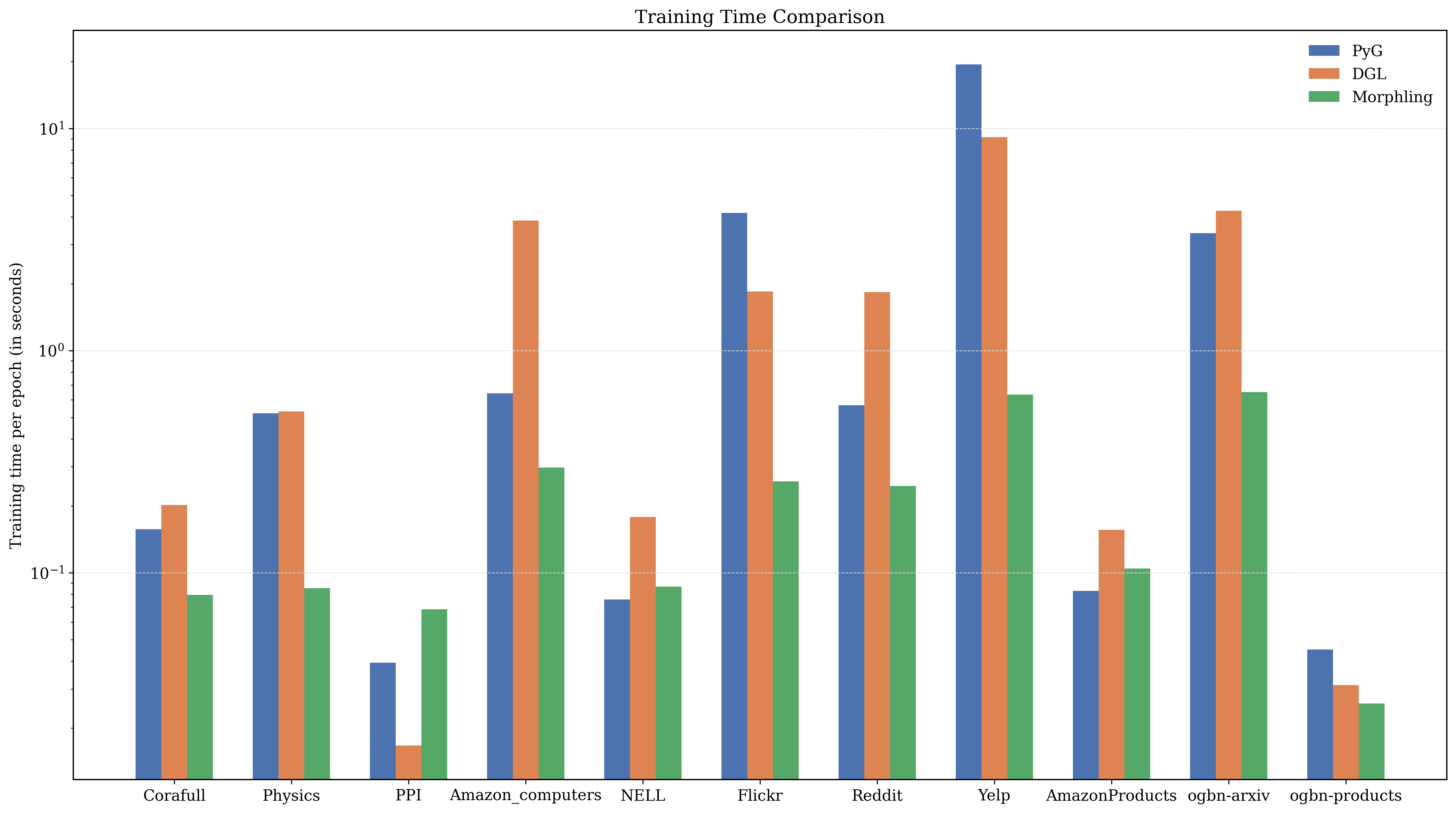}
    \caption{Per-epoch training time of Morphling, PyG, and DGL on MPI}
    \label{fig:mpitraining}
\end{figure}

We evaluate Morphling in a multi-node distributed setting and compare its performance with \texttt{pytorch\_geometric.distributed} (PyG) and DGL. Figure~\ref{fig:mpispeedup} reports relative speedups, and Figure~\ref{fig:mpitraining} presents per-epoch latencies.

\subsubsection{Throughput Analysis}
Morphling is optimized for massive, communication-intensive graphs, and its advantages become most pronounced at scale. On \textit{AmazonProducts}, for example, PyG and DGL require \textbf{3880.4 s} and \textbf{1824.4 s} per epoch, respectively. In contrast, Morphling completes an epoch in \textbf{126.9 s}, yielding a \textbf{30.58$\times$} speedup over PyG and \textbf{14.38$\times$} over DGL. Similarly, on \textit{Reddit}, Morphling reduces epoch time to \textbf{51.53 s}, achieving speedups of \textbf{16.19$\times$} (vs.\ PyG) and \textbf{7.16$\times$} (vs.\ DGL).

On medium-sized graphs (e.g., NELL, Yelp), Morphling maintains 2--13$\times$ gains over both frameworks. However, on small datasets (e.g., PPI, Flickr), the fixed overhead of the distributed runtime dominates the short computation time, resulting in parity or slight performance regression compared to DGL's fused kernels. Averaged across all benchmarks, Morphling achieves a \textbf{6.22$\times$} speedup over PyG and \textbf{5.65$\times$} over DGL.

\subsubsection{Architectural Attribution}
Morphling’s distributed scalability stems from two design choices:  
(1) A \textbf{Degree-Aware Hierarchical Partitioner} that balances computational load ($\sum \text{deg}(v)$) rather than just vertex counts, preventing straggler ranks on power-law graphs.  
(2) A \textbf{Pipelined Runtime} that masks synchronization latency. By overlapping halo exchange with forward-pass preparation and overlapping weight gradient reduction with backward-pass computation, Morphling effectively hides the communication costs that bottleneck standard synchronous implementations.

\subsection{Memory Behavior Analysis}
\label{sec:mem_analysis}

Table~\ref{tab:memoryusage} summarizes peak system memory consumption. Morphling consistently maintains the smallest footprint, enabling full‐graph training even on datasets where standard frameworks exceed hardware capacity.

\subsubsection{Capacity and Scalability}
The most critical disparity arises on \textit{AmazonProducts}. Morphling trains on the full graph using only \textbf{9.03 GB}. In contrast, PyG exceeds the 192~GB system limit and fails to initialize. Even when the dataset is downsampled to 75\%, PyG still consumes \textbf{140.24 GB}. DGL successfully runs the full graph but requires \textbf{25.10 GB}, nearly \textbf{2.8$\times$} more than Morphling. Similar trends on \textit{Reddit} and \textit{ogbn-products} demonstrate that Morphling's memory efficiency remains stable as graph size increases.

\subsubsection{Complexity Analysis}
The memory gap is structural and follows directly from the execution models. PyG’s \texttt{gather–scatter} paradigm materializes source and destination feature tensors of size $[|\mathcal{E}| \times F]$, producing a peak memory cost of:
\begin{equation}
    \mathcal{M}_{\text{PyG}} \approx \mathcal{O}(|\mathcal{E}| \times F) + \mathcal{O}(|\mathcal{V}| \times F).
\end{equation}
On dense graphs, the $|\mathcal{E}| \times F$ term dominates and leads to the observed OOM failures.

Morphling avoids edge‐tensor materialization entirely. Its fused C++ kernels accumulate messages directly into node embeddings without constructing intermediate $[|\mathcal{E}|\times F]$ buffers, bounding memory usage by:
\begin{equation}
    \mathcal{M}_{\text{Morphling}} \approx \mathcal{O}(|\mathcal{V}| \times F).
\end{equation}
This yields a reduction factor that grows proportionally with the graph’s average degree—up to \textbf{15.5$\times$} on dense graphs such as \textit{AmazonProducts}—while maintaining near parity on smaller or moderate-density datasets like \textit{ogbn-arxiv}. DGL falls between the two extremes due to its use of multiple adjacency formats (CSR/CSC) and per-layer message buffers, producing a memory footprint larger than Morphling's but lower than PyG's.

 \begin{figure}[htbp]
    \raggedright
    \includegraphics[width=0.5\textwidth]{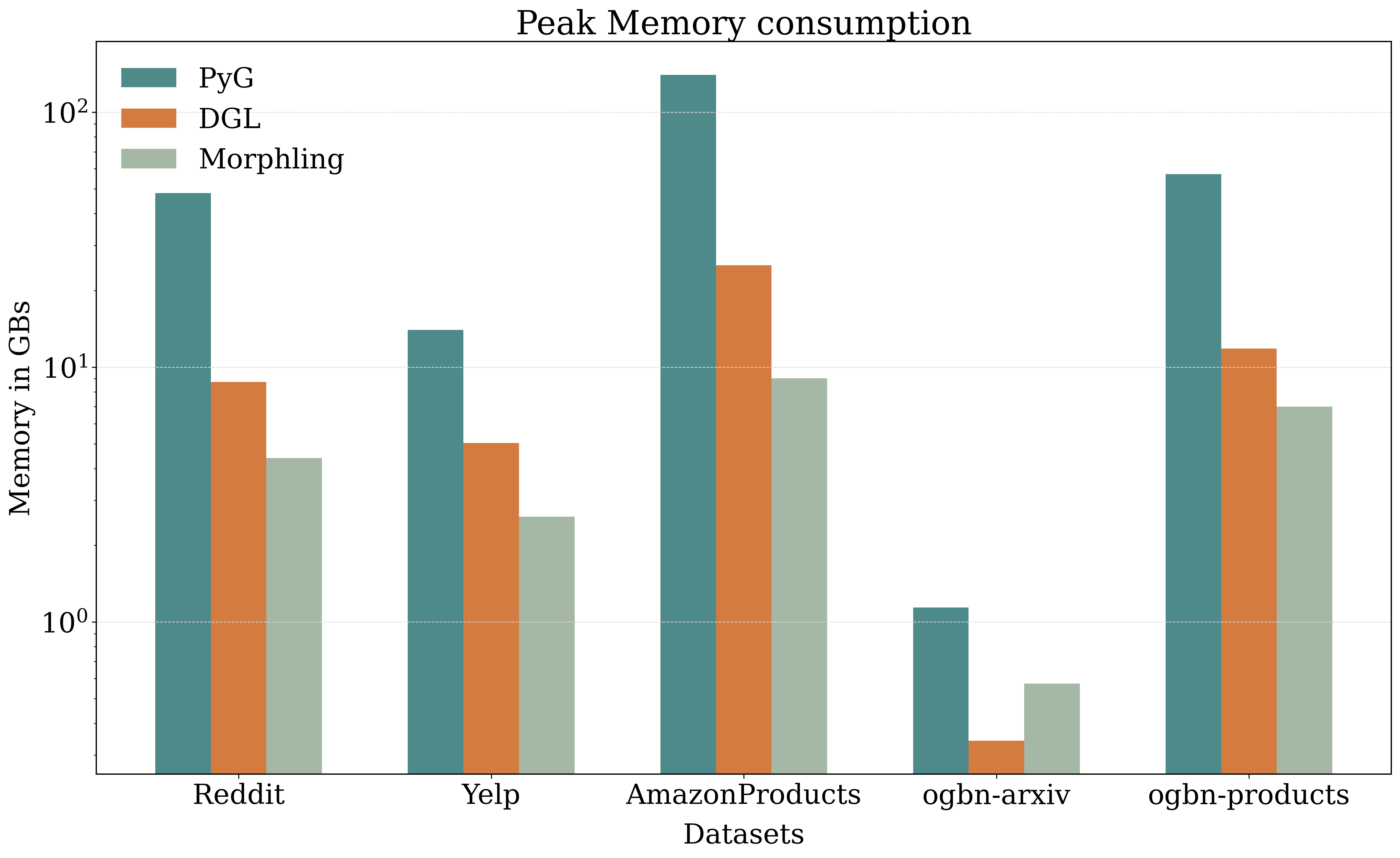}
    \caption{Peak memory comparison between Pyg, DGL, and Morphling}
    \label{fig:peakmem}
\end{figure}

Morphling avoids allocating such edge-sized feature buffers. Instead, it operates directly on node features without materializing per-edge feature tensors, which eliminates the dominant memory term present in PyG. This design enables Morphling to maintain low peak memory usage even on graphs with tens or hundreds of millions of edges, as reflected in the
dramatic savings observed across the evaluated datasets.
\section{Conclusion}
\label{sec:conclusion}

Existing GNN frameworks face a persistent mismatch between the irregular, sparse structure of graph workloads and the dense, uniform execution models of modern accelerators. We presented \textbf{Morphling}, a cross-platform synthesis framework that addresses this gap through architecture-aware specialization. By unifying a sparsity-aware execution engine, fused kernel synthesis, and a pipelined distributed runtime, Morphling substantially reduces the memory overheads and synchronization costs that constrain state-of-the-art systems.

Our evaluation shows that these design choices translate into decisive performance gains. On single-node CPUs and GPUs, Morphling improves training throughput by an average of \textbf{20.21$\times$} and \textbf{15.48$\times$}, respectively, with peak speedups reaching \textbf{66$\times$} on sparse workloads. Its memory-efficient kernel design reduces peak footprint by up to \textbf{15.5$\times$}, enabling full-graph training on large datasets such as \textit{AmazonProducts} within typical cluster node memory budgets—where PyG and DGL either fail or fall back to subsampled execution. In distributed settings, Morphling’s degree-balanced partitioner and split-phase communication pipeline scale effectively across nodes, outperforming PyG by up to \textbf{30$\times$}. Together, these results position Morphling as a robust foundation for next-generation, scale-out graph learning systems.

\section{Future Work}
\label{sec:future_work}

The future extensions of Morphling span three directions. First, deepening the compiler stack with \textbf{automated schedule tuning} and \textbf{cross-layer operator fusion} will broaden support for emerging architectures such as attention-based GNNs. Second, enabling \textbf{mixed-precision} and quantized execution (FP16/BF16/INT8) can further increase throughput and reduce memory use on modern Tensor Core accelerators. Finally, expanding backend coverage to \textbf{heterogeneous accelerators} (FPGAs, IPUs) and augmenting the distributed runtime with \textbf{topology-aware communication} and adaptive partitioning will improve performance on large-scale, hierarchical cluster environments.




\begin{thebibliography}{1}

\bibitem{pyg}
M.~Fey and J.~E. Lenssen,
``Fast Graph Representation Learning with PyTorch Geometric,''
in \emph{Proc.\ Workshop on Representation Learning on Graphs and Manifolds at ICLR}, 2019.   \url{https://arxiv.org/abs/1903.02428}


\bibitem{dgl}
M.~Wang, D.~Zheng, Z.~Ye, Q.~Gan, M.~Li, X.~Song, J.~Zhou, C.~Ma, L.~Yu, Y.~Gai, T.~Xiao, T.~He, G.~Karypis, J.~Li, and Z.~Zhang,
``Deep Graph Library: A Graph-Centric, Highly-Performant Package for Graph Neural Networks,''
in \emph{arXiv preprint arXiv:1909.01315}, Sept. 2019.   \url{https://arxiv.org/abs/1909.01315}


\bibitem{gnnadvisor}
Y.~Wang, B.~Feng, G.~Li, S.~Li, L.~Deng, Y.~Xie, and Y.~Ding,
``GNNAdvisor: An Adaptive and Efficient Runtime System for GNN Acceleration on GPUs,''
in \emph{Proc.\ 15th USENIX Symposium on Operating Systems Design and Implementation (OSDI)}, 2021, pp.~515–531.   \url{https://www.usenix.org/conference/osdi21/presentation/wang-yuke}


\bibitem{dgcl}
H.~Zhang, Z.~Yu, G.~Dai, G.~Huang, Y.~Ding, Y.~Xie, and Y.~Wang,
``Understanding GNN Computational Graph: A Coordinated Computation, IO, and Memory Perspective,''
in \emph{Proc.\ 5th Machine Learning and Systems Conference (MLSys)}, 2022, pp.~467--484.   \url{https://arxiv.org/abs/2110.09524}

\bibitem{aligraph}
R.~Zhu, K.~Zhao, H.~Yang, W.~Lin, C.~Zhou, B.~Ai, Y.~Li, and J.~Zhou,
``AliGraph: A Comprehensive Graph Neural Network Platform,''
in \emph{Proc.\ The 12th International Conference on Very Large Data Bases (VLDB)}, vol.~12, no.~12, 2019, pp.~2094–2105. doi:10.14778/3352063.3352127.   \url{https://doi.org/10.14778/3352063.3352127}


\bibitem{euler}
F.~Huang, W.~Zhang, K.~Zhai, J.~Zhu, R.~Chen, W.~Wu, and X.~Wan,
``Euler: A System for Large-Scale Graph Analytics,''
in \emph{Proc.\ 24th ACM SIGKDD International Conference on Knowledge Discovery \& Data Mining (KDD)}, 2018, pp.~316–324. doi:10.1145/3219819.3220063.   \url{https://doi.org/10.1145/3219819.3220063}


\bibitem{distdgl}
D.~Zheng, C.~Ma, M.~Wang, J.~Zhou, Q.~Su, X.~Song, Q.~Gan, Z.~Zhang, and G.~Karypis,
``DistDGL: Distributed Graph Neural Network Training for Billion-Scale Graphs,''
in \emph{10th IEEE/ACM Workshop on Irregular Applications: Architectures and Algorithms (IA\(^3\))}, 2020, pp.~36–44.   \url{https://arxiv.org/abs/2010.05337}


\bibitem{graphit}
Y.~Zhang, A.~Brahmakshatriya, X.~Chen, L.~Dhulipala, S.~Kamil, S.~Amarasinghe, and J.~Shun,
``GraphIt: A High-Performance DSL for Graph Analytics,''
in \emph{Proc.\ 33rd ACM SIGPLAN Conference on Programming Language Design and Implementation (PLDI)}, 2018. doi:10.1145/3276491.   \url{https://doi.org/10.1145/3276491}


\bibitem{ligra}
J.~Shun and G.~E. Blelloch,
``Ligra: A Lightweight Graph Processing Framework for Shared Memory,''
in \emph{Proc.\ 18th ACM SIGPLAN Symposium on Principles and Practice of Parallel Programming (PPoPP)}, Feb. 23-27 2013, Shenzhen, China, pp.~135–146. doi:10.1145/2517327.2442530.


\bibitem{galois}
M.~Kulkarni, K.~Pingali, B.~Walter, K.~Tung, and J.~Hollingsworth,
``The Galois System: Optimistic Parallelization of Irregular Programs,''
in \emph{Proc.\ 20th IEEE International Parallel \& Distributed Processing Symposium (IPDPS)}, 2008, pp.~1-12.   \url{https://ecommons.cornell.edu/bitstreams/707e6678-bb5b-4dc6-b590-1b58e409d802/download}


\bibitem{gemini}
X.~Zhu, W.~Chen, W.~Zheng, and X.~Ma,
``Gemini: A Computation-Centric Distributed Graph Processing System,''
in \emph{Proc.\ 12th USENIX Symposium on Operating Systems Design and Implementation (OSDI)}, Savannah, GA, USA, Nov. 2016, pp.~301–316.   \url{https://www.usenix.org/conference/osdi16/technical-sessions/presentation/zhu}


\bibitem{starplat}
N.~Behera, A.~Kumar, E.~Rajadurai T., S.~Nitish, R.~Pandian M., and R.~Nasre,
``StarPlat: A Versatile DSL for Graph Analytics,''
\emph{Journal of Parallel and Distributed Computing}, vol.~183, 2024, Article 104967. doi:10.1016/j.jpdc.2024.104967.  \url{https://arxiv.org/abs/2305.03317}


\bibitem{tensorflow}
M.~Abadi, P.~Barham, J.~Chen, Z.~Chen, A.~Davis, J.~Dean, M.~Devin, S.~Ghemawat,
G.~Irving, M.~Isard, M.~Kudlur, J.~Levenberg, R.~Monga, S.~Moore, D.~Murray, B.~Steiner,
P.~Tucker, V.~Vasudevan, P.~Warden, M.~Wicke, Y.~Yu, and X.~Zheng,
``TensorFlow: A System for Large-Scale Machine Learning,''
in \emph{Proc.\ 12th USENIX Symposium on Operating Systems Design and Implementation (OSDI)}, 
Savannah, GA, USA, Nov. 2016, pp.~265–283.  \url{https://www.usenix.org/system/files/conference/osdi16/osdi16-abadi.pdf}


\bibitem{pytorch}
A.~Paszke, S.~Gross, F.~Massa, A.~Lerer, J.~Bradbury, G.~Chanan, T.~Killeen, Z.~Lin, N.~Gimelshein, L.~Antiga, A.~Desmaison, A.~Köpf, E.~Yang, Z.~DeVito, M.~Raison, A.~Tejani, S.~Chilamkurthy, B.~Steiner, L.~Fang, J.~Bai, and S.~Chintala,
``PyTorch: An Imperative Style, High-Performance Deep Learning Library,''
in \emph{Proc.\ 33rd International Conference on Neural Information Processing Systems (NeurIPS)}, Vancouver, Canada, 2019, pp.~8024–8035.  \url{https://arxiv.org/abs/1912.01703}




\bibitem{scalefree}
H.~Guo, W.~Guo, Y.~Gao, R.~Tang, X.~He, and W.~Liu,
``ScaleFreeCTR: MixCache-based Distributed Training System for CTR Models with Huge Embedding Table,''
in \emph{Proc.\ 44th International ACM SIGIR Conference on Research and Development in Information Retrieval (SIGIR ’21)}, July 11-15 2021, Virtual Event, Canada, 10 pages. doi:10.1145/3404835.3462976.  \url{https://arxiv.org/abs/2104.08542}


\bibitem{seastar}
Y.~Wu, K.~Ma, Z.~Cai, T.~Jin, B.~Li, C.~Zheng, J.~Cheng, and F.~Yu,
``Seastar: Vertex-Centric Programming for Graph Neural Networks,''
in \emph{Proc.\ ACM / IEEE European Conference on Computer Systems (EuroSys ’21)}, Apr. 26-29 2021, 17 pages. doi:10.1145/3447786.3456247.   \url{https://doi.org/10.1145/3447786.3456247}


\bibitem{featgraph}
Y.~Hu, Z.~Ye, M.~Wang, J.~Yu, D.~Zheng, M.~Li, Z.~Zhang, Z.~Zhang, and Y.~Wang,
``FeatGraph: A Flexible and Efficient Backend for Graph Neural Network Systems,''
in \emph{Proc.\ IEEE/ACM International Conference for High Performance Computing, Networking, Storage and Analysis (SC)}, 2020.   \url{https://arxiv.org/abs/2008.11359}


\bibitem{halide}
J.~Ragan-Kelley, A.~Adams, S.~Paris, M.~Levoy, and F.~Durand,
``Halide: A Language and Compiler for Optimizing Parallelism, Locality, and Recomputation in Image Processing Pipelines,''
in \emph{Proc.\ ACM SIGPLAN Conference on Programming Language Design and Implementation (PLDI)}, 
2013, pp.~519--530.   \url{https://dl.acm.org/doi/10.1145/2491956.2462176}


\bibitem{tvm}
T.~Chen, T.~Moreau, Z.~Jiang, L.~Ceze, C.~Guestrin, and A.~Krishnamurthy,
``TVM: An Automated End-to-End Optimizing Compiler for Deep Learning,''
in \emph{Proc.\ 13th USENIX Symposium on Operating Systems Design and Implementation (OSDI)}, 
2018, pp.~578--594.   \url{https://www.usenix.org/conference/osdi18/presentation/chen}


\bibitem{taco}
F.~Kjolstad, S.~Kamil, S.~Chou, D.~Lugato, and S.~Amarasinghe,
``The Tensor Algebra Compiler,''
in \emph{Proc.\ ACM Program.\ Lang.}, vol.\ 1, no.\ OOPSLA, Article 77, Oct.\ 2017. doi:10.1145/3133901.   \url{https://doi.org/10.1145/3133901}


\bibitem{tiramisu}
R.~Baghdadi, Y.~Zou, M.~Shirinzadeh, P.~Zhao, M.~Ripeanu, T.~M.~Smith, A.~Sanan, 
R.~Mullapudi, J.~Ragan-Kelley, and S.~Amarasinghe,
``Tiramisu: A Polyhedral Compiler for Expressing Fast and Portable Code,''
in \emph{Proc.\ IEEE/ACM International Symposium on Code Generation and Optimization (CGO)}, 
2019, pp.~193--205. \url{https://doi.org/10.1109/CGO.2019.8661197}


\bibitem{tc}
N.~Vasilache, O.~Zinenko, T.~Theodoridis, P.~Goyal, Z.~DeVito, W.~S. Moses, S.~Verdoolaege, A.~Adams, and A.~Cohen,
``Tensor Comprehensions: Framework-Agnostic High-Performance Machine Learning Abstractions,''
in \emph{Proc.\ of the 2nd ACM/IEEE Workshop on Machine Learning and Systems (MLSys) — Pre-print / ArXiv}, 2018.   \url{https://arxiv.org/abs/1802.04730}


\bibitem{kokkos}
H.~C. Edwards, C.~R. Trott, and D.~Sunderland,
``Kokkos: Enabling manycore performance portability through polymorphic memory access patterns,''
\emph{Journal of Parallel and Distributed Computing}, vol.~74, no.~12, pp.~3202--3216, Dec. 2014. doi:10.1016/j.jpdc.2014.07.003.   \url{https://doi.org/10.1016/j.jpdc.2014.07.003}

\bibitem{raja}
D.~A. Beckingsale, J.~Burmark, R.~Hornung, H.~Jones, W.~Killian, A.~J. Kunen, O.~Pearce, P.~Robinson, B.~S. Ryujin, and T.~R. W. Scogland,
``RAJA: Portable Performance for Large-Scale Scientific Applications,''
in \emph{Proc. IEEE/ACM International Workshop on Performance, Portability and Productivity in HPC (P3HPC)}, 2019, pp. 71-81. \url{https://www.osti.gov/servlets/purl/1573949}



\bibitem{sycl}
R.~Reyes and V.~Lomüller,
``SYCL: Single-source C++ accelerator programming,''
in \emph{Advances in Parallel Computing, Vol. 27: Parallel Computing — On the Road to Exascale}, 
R.~T.~Morris, Ed., IOS Press, 2016, pp. 673-682. \url{https://doi.org/10.3233/978-1-61499-621-7-673}


\bibitem{oneapi}
Intel Corporation,
``Intel® oneAPI Programming Guide,’’
Intel Corporation, 2025. \href{https://www.intel.com/content/www/us/en/docs/oneapi/programming-guide/2025-0/overview.html}{oneAPI Documentation}


\bibitem{opencl}
J.~E. Stone, D.~Gohara, and G.~Shi,
``OpenCL: A parallel programming standard for heterogeneous computing systems,''
\emph{Computing in Science \& Engineering}, 
vol.~12, no.~3, pp.~66--73, 2010. \url{https://doi.org/10.1109/MCSE.2010.69}




\bibitem{gcn}
T.~N. Kipf and M.~Welling,
``Semi-supervised classification with graph convolutional networks,''
in \emph{Proc. Int. Conf. Learn. Represent. (ICLR)}, 
2017. \url{https://arxiv.org/abs/1609.02907}


\bibitem{graphsage}
W.~L. Hamilton, R.~Ying, and J.~Leskovec, 
``Inductive representation learning on large graphs,'' 
in \emph{Proc. Adv. Neural Inf. Process. Syst. (NeurIPS)}, 
vol.~30, 2017. \url{https://arxiv.org/abs/1706.02216}


\bibitem{gin}
K.~Xu, W.~Hu, J.~Leskovec, and S.~Jegelka,
``How powerful are graph neural networks?,'' 
in \emph{Proc. Int. Conf. Learn. Represent. (ICLR)}, 2019. \url{https://arxiv.org/abs/1810.00826}


\bibitem{gat}
P.~Veličković, G.~Cucurull, A.~Casanova, A.~Romero, P.~Liò, and Y.~Bengio, 
``Graph attention networks,'' in \emph{Proc. Int. Conf. Learn. Represent. (ICLR)}, 
2018. \url{https://arxiv.org/abs/1710.10903}


\bibitem{metis}
G.~Karypis and V.~Kumar, ``A fast and high quality multilevel scheme for partitioning irregular graphs,'' 
\emph{SIAM Journal on Scientific Computing}, vol.~20, no.~1, pp.~359--392, 1998. 
doi: \href{https://doi.org/10.1137/S1064827595287997}{10.1137/S1064827595287997}.


\bibitem{pyg_sparse_tensors}
PyTorch Geometric, ``Sparse tensor support,'' 
\emph{PyTorch Geometric Documentation}.   \href{ https://pytorch-geometric.readthedocs.io/en/latest/advanced/sparse_tensor.html}{Pytorch\_Geometric Documentation}




\end{thebibliography}
\end{document}